\pdfoutput=1

\documentclass[11pt]{article}

\usepackage[preprint]{acl}

\usepackage{times}
\usepackage{latexsym}

\usepackage[T1]{fontenc}

\usepackage[utf8]{inputenc}

\usepackage{microtype}

\usepackage{inconsolata}

\usepackage{graphicx}

\usepackage{adjustbox}
\usepackage{bbding}
\usepackage{pifont}
\usepackage{microtype}
\usepackage{hyperref}
\usepackage{url}
\usepackage{booktabs}
\usepackage{multirow}
\usepackage[normalem]{ulem}
\usepackage{graphicx}
\usepackage{amssymb}
\usepackage{makecell}
\usepackage{amsmath}
\usepackage{subfigure} 
\usepackage[most]{tcolorbox}
 \usepackage{soul}
\usepackage{array}       
\usepackage{tabularx}    
\usepackage{xcolor}    
\usepackage{colortbl}

\usepackage{longtable}

\definecolor{yellow}{HTML}{F6BD60}
\definecolor{white}{HTML}{FFE0C1}
\definecolor{pink}{HTML}{F5CAC3}
\definecolor{tale}{HTML}{84A59D}
\definecolor{red}{HTML}{F28080}
\definecolor{orange}{HTML}{FF7F00}
\definecolor{green1}{HTML}{72C3A3}
\definecolor{green2}{HTML}{70B48F}
\definecolor{midgreen}{HTML}{39ad6e} 
\definecolor{orange}{HTML}{FE8019}
\definecolor{grey}{HTML}{EBDBB2}
\definecolor{light_grey}{HTML}{dde4e5}
\definecolor{brain}{HTML}{FFABBE}
\definecolor{blue}{HTML}{076678}
\definecolor{purple}{HTML}{5861AC}
\definecolor{narrative}{HTML}{458588}
\definecolor{white2}{HTML}{F8F5E9}
\definecolor{tablewhite}{HTML}{E5E1DA}

\newcolumntype{L}{>{\raggedright\arraybackslash}X} 
\newcolumntype{C}{>{\centering\arraybackslash}X}

\title{Evaluating LLMs' Assessment of Mixed-Context Hallucination\\ Through the Lens of Summarization}

\author{
    Siya Qi$^{\clubsuit}$ \quad 
    Rui Cao$^{\spadesuit}$ \quad 
    Yulan He$^{\clubsuit,\diamondsuit}$ \quad
    Zheng Yuan$^{\clubsuit}$\\
    $^\clubsuit$King's College London \quad
    $^\spadesuit$Cambridge University \\
    $^\diamondsuit$The Alan Turing Institute \\
    {\small \tt \{siya.qi, zheng.yuan, yulan.he\}@kcl.ac.uk}  \\[-0.4em]
    {\small \tt rc990@cam.ac.uk} 
    \vspace{-1em}
}

\begin{document}
\maketitle
\begin{abstract}
With the rapid development of large language models (LLMs), LLM-as-a-judge has emerged as a widely adopted approach for text quality evaluation, including hallucination evaluation. While previous studies have focused exclusively on single-context evaluation (e.g., discourse faithfulness or world factuality), real-world hallucinations typically involve mixed contexts, which remains inadequately evaluated.
In this study, we use summarization as a representative task to comprehensively evaluate LLMs' capability in detecting mixed-context hallucinations, specifically distinguishing between factual and non-factual hallucinations. Through extensive experiments across direct generation and retrieval-based models of varying scales, our main observations are: 
(1) LLMs' intrinsic knowledge introduces inherent biases in hallucination evaluation;
(2) These biases particularly impact the detection of factual hallucinations, yielding a significant performance bottleneck;
(3) The fundamental challenge lies in effective knowledge utilization, balancing between LLMs' intrinsic knowledge and external context for accurate mixed-context hallucination evaluation.

\end{abstract}

\section{Introduction}

\begin{figure}[htbp]
    \centering
    \includegraphics[width=0.48\textwidth]{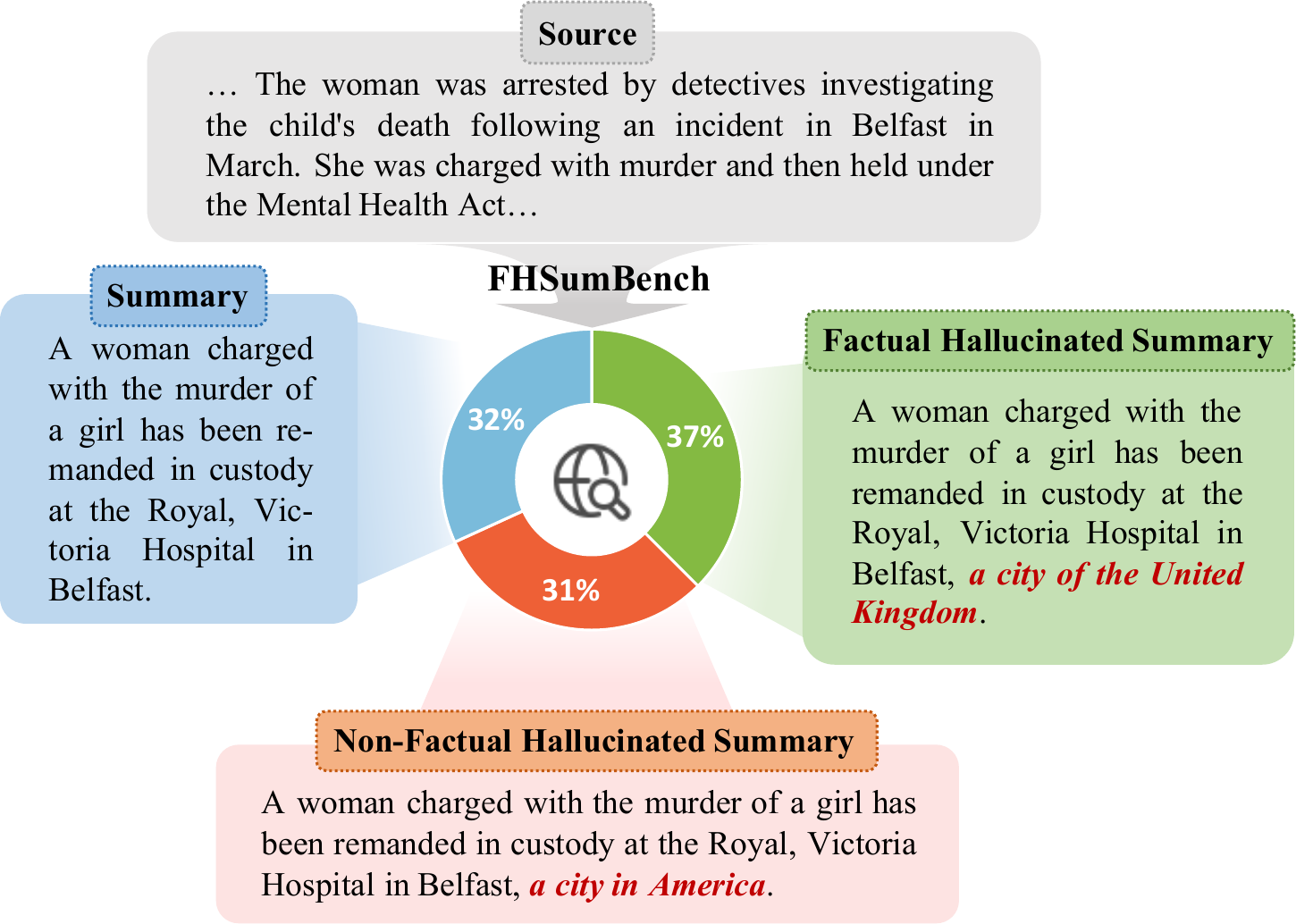}
    \caption{Examples of our automated construction for mixed-context hallucination datasets, where "a city of the United Kingdom" is the correct description of "Belfast", constructed as factual hallucination. "A city in America" is an incorrect description of "Belfast", constructed as a non-factual hallucination.}
    \label{fig:exp1}
\end{figure}

Large language models (LLMs) generate coherent text and follow instructions across diverse tasks~\cite{zhao2024surveylargelanguagemodels, minaee2024largelanguagemodelssurvey}.
However, a critical challenge in scaling LLM applications is hallucination, where the generated content lacks factual grounding~\cite{lin-etal-2022-truthfulqa, li2023halueval} or deviates from the intended discourse context~\cite{zhou-etal-2023-context}.
This issue makes hallucination evaluation essential for improving LLM reliability and developing more robust models~\cite{huang_survey_2023, zhang2023siren, ji2023survey, tonmoy_comprehensive_2024}. 
Recent studies have explored the potential of using LLMs themselves to assess hallucinations, reporting strong performance in hallucination detection with LLM-based evaluators~\cite{gu2025surveyllmasajudge, min-etal-2023-factscore, chern2023factool}.

In practical applications, hallucinations typically derive from two primary sources: one originates from the discourse context, affecting faithfulness; while the other arises from external knowledge or the model's inherent knowledge, influencing factuality~\cite{maynez-etal-2020-faithfulness}. Previous LLM-based evaluators for hallucinations have focused on assessing either faithfulness~\cite{manakul-etal-2023-selfcheckgpt, wu-etal-2023-wecheck, zha2023alignscore} or factuality~\cite{chern2023factool,min-etal-2023-factscore,dhuliawala-etal-2024-chain, hu2024refcheckerreferencebasedfinegrainedhallucination} of the text under evaluation. However, hallucinations can emerge from a combination of discourse context and external information -- a mixed-context setting that remains under-explored in LLM-based evaluation.
Many real-world scenarios operate within this mixed-context setting.
For instance, in summarization, a system might fabricate details while still presenting factually correct information; in customer support chatbots, it may hallucinate from chat histories alongside accurate product information; and in image captioning, models might provide factually correct descriptions about non-existing objects in an image.
Evaluating these mixed-context hallucinations requires simultaneous assessment of both faithfulness and factuality, which presents unique challenges for developing robust evaluation metrics and benchmarks.
In this work, we focus on the summarization task to assess LLMs' capability under this setting.

With mixed-context considered, hallucinations in summarization
can be categorized into two types\footnote{We focus on the news summarization task, where source articles are assumed to be factually correct, so a hallucinated summary cannot be both faithful and non-factual.}:
\textbf{factual hallucinations}, where the generated content is factually correct but inconsistent with the source (e.g., the green block in Figure~\ref{fig:exp1}); and \textbf{non-factual hallucinations}, where the output contains outright factual inaccuracies (e.g., the red block in Figure~\ref{fig:exp1}).
Existing datasets for mixed-context hallucinations in summarization ~\cite{maynez-etal-2020-faithfulness, cao-etal-2022-hallucinated, dong-etal-2022-faithful} are limited both in scale and exhibit unbalanced distributions of hallucination instances, making them inadequate to assess LLMs' capability in handling mixed-context hallucinations.
To address this limitation,
this study introduces the first automated pipeline for constructing mixed-context hallucination evaluation datasets, and aims to investigate the following research questions:
\textbf{RQ1)} How do LLM-based hallucination evaluators perform in the mixed-context setting with different prompting strategies? 
Given the rapid advancement of retrieval-augmented generation (RAG)~\cite{gao_retrieval-augmented_2024}, we then examine \textbf{RQ2)} How can LLMs better leverage external knowledge to detect hallucinations?
Among the hallucination categories, we ask \textbf{RQ3)} Which category benefits the most across evaluation methods and models?
Finally, a fundamental question in LLM research, \textbf{RQ4)} Does scaling up model size lead to better hallucination assessment?
These questions collectively address the challenges in mixed-context hallucination assessment, with our findings laying the groundwork for scalable LLM self-evaluation and self-evolution.
The contributions and key observations of this work are as follows:

\begin{enumerate}
    \item We comprehensively evaluate LLM-as-a-judge approaches for mixed-context hallucination detection through summarization, introducing an effective, well-balanced, and easily scalable benchmark, FHSumBench\footnote{Data and code are available at \url{https://github.com/cece00/FHSumBench}}.
    
    \item Our extensive experiments across direct generation and retrieval-based LLMs of varying scales reveal that model scaling does not guarantee better performance. However, the main performance bottleneck lies in the detection of factual hallucinations.
    
    \item We identify that effective knowledge utilization remains the primary challenge in mixed-context hallucination evaluation. 
    Prompt engineering yields greater improvements on smaller LLMs. While external knowledge augmentation enhances accuracy, it requires carefully designed retrieval strategies.

\end{enumerate}

\section{Related work}
\label{sec:related_work}

\paragraph{Mixed-Context Hallucination}

Context plays a crucial role in hallucination evaluation. \citet{hu2024refcheckerreferencebasedfinegrainedhallucination} examine the hallucination detection capabilities of LLMs in the question-answer task under different contextual settings. However, their evaluation is limited to isolated contexts.
For summarization, while the source document serves as a context input, the factuality of the summary is equally critical.
~\citet{maynez-etal-2020-faithfulness} first systematically annotated the faithfulness and factuality separately on the XSum~\cite{narayan-etal-2018-dont} dataset, using summaries generated by small language models. Later, ~\citet{cao-etal-2022-hallucinated} introduced another dataset with entity-level annotations, though it remained limited to XSum. The annotation process demands extensive effort from domain experts and heavily relies on annotator agreement, making it both time-consuming and resource-intensive.

\paragraph{Hallucination Evaluation Methods}
Most previous works only focus on one aspect of hallucination evaluation, which is either faithfulness or factuality. For faithfulness evaluation, a natural approach is to assess the entailment between the text and the context document~\cite{wu-etal-2023-wecheck, zha2023alignscore, goyal-durrett-2020-evaluating} or to extract and compare answers by querying them~\cite{manakul-etal-2023-mqag, durmus-etal-2020-feqa, fabbri-etal-2022-qafacteval}. For factuality evaluation, LLMs intrinsic knowledge~\cite{chen-etal-2023-beyond, huang2024ufo} and external knowledge source~\cite{min-etal-2023-factscore, chern2023factool, dhuliawala-etal-2024-chain} are both used for detection.
With both aspects considered, 
EntFA~\cite{cao-etal-2022-hallucinated} utilizes entity original and context-conditioned features to detect factual and non-factual hallucinations.

\section{FHSumBench}
\label{sec:bench}
\begin{figure}[tbp]
    \centering
    \includegraphics[width=0.45\textwidth]{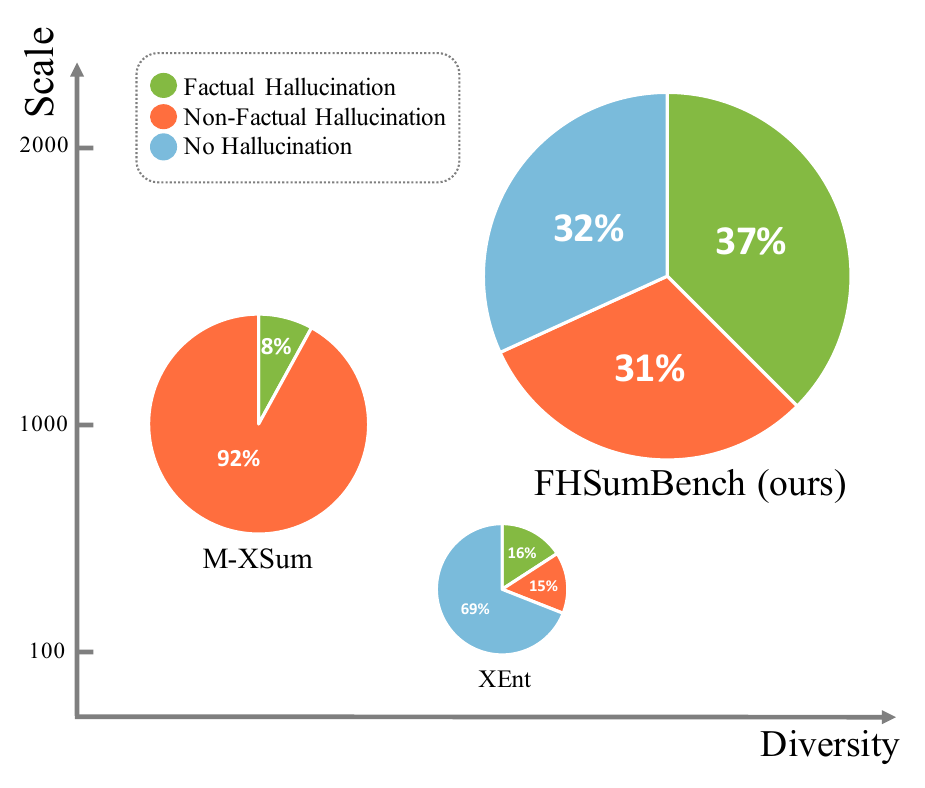}
    \caption{The size and distributions of FHSumBench, M-XSum, and XEnt datasets.}
    \label{fig:data_pie}
\end{figure}

To evaluate models' ability to detect mixed-context hallucinations in summaries, an appropriate test set is essential. 
To the best of our knowledge, the XEnt dataset~\cite{cao-etal-2022-hallucinated} is the only one providing factual hallucination annotations for summaries. However, among the 240 samples in the test set, only 29\% of entities are annotated as factual or non-factual hallucinations, which is insufficient for assessing LLMs.
Another related dataset is M-XSum~\cite{maynez-etal-2020-faithfulness}, which is also built on the XSum dataset and includes annotations for separate labels: faithfulness and factuality. The faithfulness annotations classify hallucinations as either intrinsic or extrinsic, 
while the factuality annotations are based on world knowledge.
After filtering out samples with inconsistent annotations, M-XSum remains an imbalanced dataset with 92\% non-factual hallucination samples, as shown in Figure~\ref{fig:data_pie}.
Therefore, to obtain more diverse evaluation data, we propose an automated pipeline capable of accurately and effectively constructing both factual and non-factual hallucinations. Using this pipeline, we construct the \textbf{Factual Hallucination in Summarization Benchmark (FHSumBench)} to evaluate the performance of existing factuality and faithfulness evaluators, as well as LLM-as-a-judge approaches.

The core component of this pipeline is \textbf{fact injection}: injecting either factual or non-factual information into correct summaries.
We use XEnt~\cite{cao-etal-2022-hallucinated} and FactCollect~\cite{feng2023factkb} as our seed datasets, which provide annotated correct summaries, defined as being both faithful and factual.\footnote{The relationship between FHSumBench and previous summarization datasets can be seen in Table~\ref{tab:dataset_compare}, Appendix~\ref{apx:dataset}.}
The first step of the pipeline is to obtain correct summaries. For XEnt dataset, we remove spans annotated as "non-factual hallucinations," while for the FactCollect dataset, we use summaries labeled as "correct."
The second step involves injecting factual information into the correct summaries. Here, we focus on entity knowledge as the basis for injection. Entities are extracted using the off-the-shelf NER tool~\cite{honnibal_spacy_2018}, and a one-sentence description of each entity is used as the injected information. 
The injected facts are extracted from Wikidata~\cite{10.1145/2629489}.
To create factual hallucinations, the correct description is appended to the corresponding entity. In contrast, non-factual hallucinations are built by appending a randomly selected, unrelated description to the entity, as in Figure~\ref{fig:exp1}.

As a result, FHSumBench contains 1,336 samples, evenly distributed across three categories of summaries, providing a balanced dataset for comprehensive evaluation.
This approach is more efficient than manual annotation, as it does not need experts or extensive time. Additionally, it is more controllable than data synthesis by LLMs, as it relies on precise descriptions for each entity. While preserving these advantages, this benchmark can effectively evaluate the ability of current LLMs and evaluators to assess factual hallucinations.

\section{LLM-Based Evaluators on Mixed-Context Hallucination}
\label{sec:method}

In this section, we describe widely used LLM-based evaluators for hallucinations. Based on whether they retrieve knowledge from external sources, we categorize the evaluators into two types: \textbf{Direct Generation Evaluators} (\textsection\ref{sec:llm-wo-retrieve}) and \textbf{Retrieval-Based Evaluators} (\textsection\ref{sec:llm-with-retrieve}).
We will evaluate the performance of these approaches by analyzing their effectiveness in assessing mixed-context hallucination.

\subsection{Direct Generation Evaluators}
\label{sec:llm-wo-retrieve}
\paragraph{Vanilla Judge}

The vanilla judge evaluates based solely on the document-summary pair. The evaluation is designed to maximize self-contained judgment, ensuring that the LLM relies solely on its intrinsic knowledge, without additional modules. Specifically, the LLM assesses faithfulness by referencing the source document, while factuality is judged based on its intrinsic knowledge.
For output $Y$ and input $X$ (comprising context $C$ and query $Q$), where $C$ includes the source document for vanilla judge, we have $P(Y \mid X)=P(Y \mid C,Q)$.

\paragraph{Prompting Strategies}

Following vanilla judge, we want to further investigate what capabilities are lacking in vanilla LLMs for objective and accurate judgment.
Specifically, we seek to determine whether these challenges arise from constraints in reasoning ability or from difficulties in comprehending the task.
To systematically analyze this, we explore two prompting strategies: in-context learning (+ICL) and chain-of-thought reasoning (+CoT). 
For \textbf{+ICL Judge}, we provide the LLM with three annotated examples, labeled for faithfulness and factuality, to facilitate in-context learning.
For \textbf{+CoT Judge}, we incorporate the phrase "Think step by step" into the prompt to guide the LLM in generating a reasoning trajectory.

\begin{table*}[htbp]
\centering
\begin{adjustbox}{width=0.75\textwidth}
\begin{tabular}{rl|rrrrrr}
\Xhline{1.5px}
\multicolumn{2}{c|}{\multirow{2}{*}{\textbf{Methods}}} & \multicolumn{3}{c}{\textbf{FHSumBench}} & \multicolumn{3}{c}{\textbf{M-XSum}} \\ 
\multicolumn{2}{c|}{}  & \multicolumn{1}{c}{\textbf{P}} & \multicolumn{1}{c}{\textbf{R}} & \multicolumn{1}{c}{\textbf{F}} & \multicolumn{1}{c}{\textbf{P}} & \multicolumn{1}{c}{\textbf{R}} & \multicolumn{1}{c}{\textbf{F}} \\ \Xhline{1.5px}
\multicolumn{2}{c|}{Random (lower bound)} & 0.3185 & 0.3184 & 0.3185 & 0.3304 & 0.2120 & 0.2583 \\
\multicolumn{2}{c|}{EntFA~\cite{cao-etal-2022-hallucinated}} & 0.3295 & 0.3119 & 0.3205 & 0.3366 & 0.3489 & 0.3426 \\ \Xhline{1.5px}
\multicolumn{1}{c}{\multirow{3}{*}{Vanilla Judge}} & Llama3-8B & 0.3652 & 0.2977 & 0.3280 & 0.3477 & 0.1914 & 0.2469 \\
\multicolumn{1}{c}{} & Qwen2.5-14B & 0.4549 & 0.4652 & 0.4600 & 0.3710 & 0.4236 & 0.3955 \\
\multicolumn{1}{c}{} & GPT-4o & \underline{0.4575} & 0.4650 & 0.4612 & \textbf{0.3905} & \underline{0.4466} & \underline{0.4166} \\ \hline
\multirow{3}{*}{+ICL} & Llama3-8B & 0.3770 & 0.3603 & 0.3685 & 0.3664 & 0.2994 & 0.3295 \\
 & Qwen2.5-14B & 0.4333 & 0.4486 & 0.4408 & \underline{0.3897} & 0.4347 & 0.4110 \\
 & GPT-4o & 0.4519 & 0.4585 & 0.4552 & 0.3806 & \textbf{0.4629} & \textbf{0.4178} \\ \hline
\multirow{3}{*}{+CoT} & Llama3-8B & 0.3437 & 0.3044 & 0.3228 & 0.3575 & 0.3021 & 0.3275 \\
 & Qwen2.5-14B & \textbf{0.4717} & \textbf{0.4749} & \textbf{0.4733} & 0.3735 & 0.3908 & 0.3819 \\
 & GPT-4o & 0.4539 & \underline{0.4739} & \underline{0.4637} & 0.3662 & 0.3766 & 0.3713 \\ \Xhline{1.5px}
\end{tabular}
\end{adjustbox}
\caption{The results using direct generation LLM evaluators on FHSumBench and M-XSum datasets. We report the precision (P), recall (R) and F1-score (F) in this table. The best and second-best results are highlighted in \textbf{bold} and \underline{underlined}, respectively. "+ICL" denotes in-context learning judge and "+CoT" denotes chain-of-thoughts judge, both of which are based on vanilla judge.}
\label{tab:result_rq1}
\end{table*}

\subsection{Retrieval-Based Evaluators}
\label{sec:llm-with-retrieve}
\begin{figure}[htbp]
    \centering
    \includegraphics[width=0.45\textwidth]{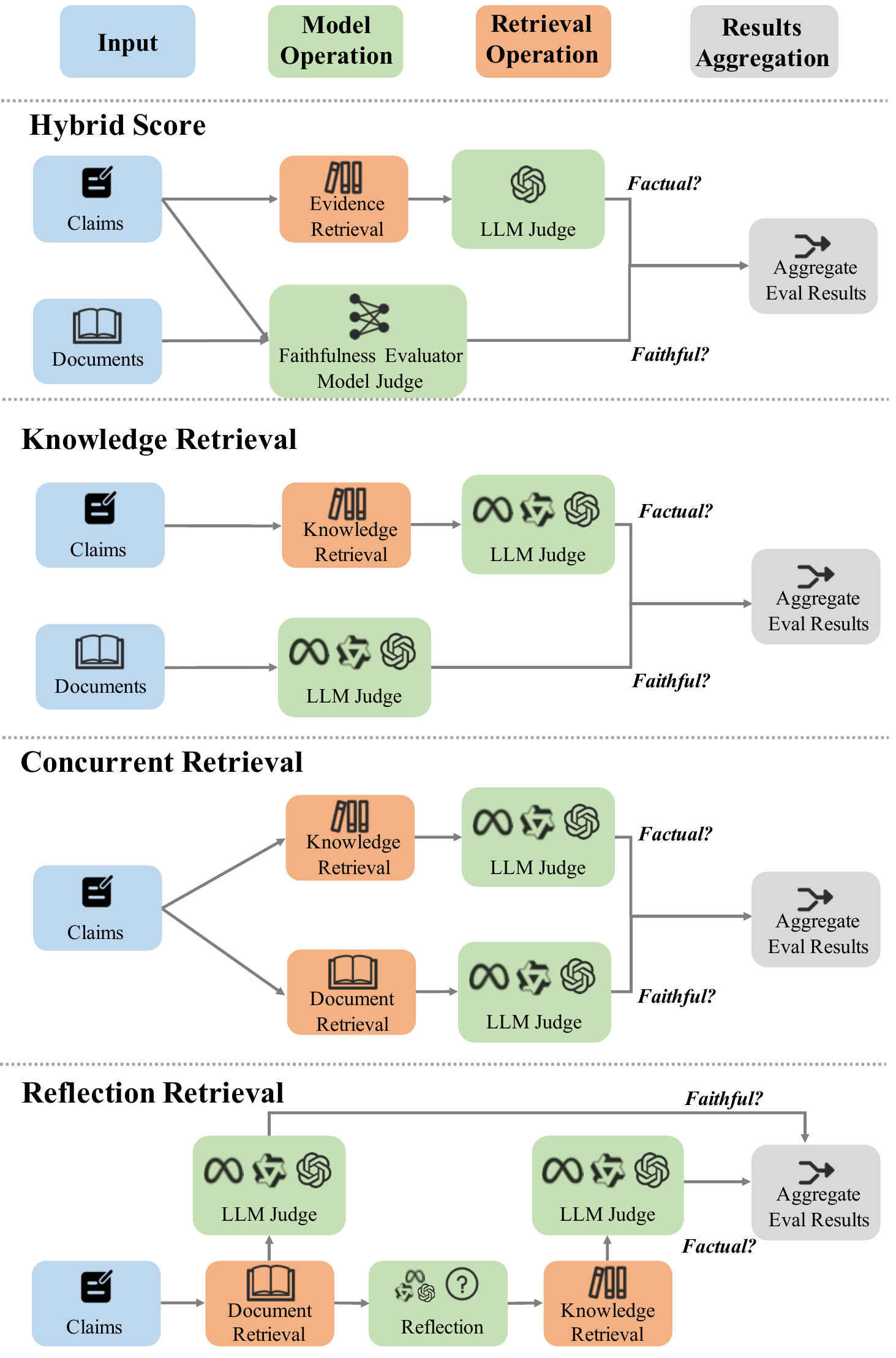}
    \caption{The pipelines of different retrieval-based evaluation methods.}
    \label{fig:retrieve}
\end{figure}

In this study, we explore evidence retrieval using previous evaluators and LLMs, comparing the following approaches to assess the faithfulness and factuality of summaries (see Figure~\ref{fig:retrieve}).

\paragraph{Hybrid Score}
We select two evaluators from different evaluation aspects for each category and combined them as baselines.
For faithfulness evaluation, we use WeCheck~\cite{wu-etal-2023-wecheck}, which aggregates results from multiple NLI models, and AlignScore~\cite{zha2023alignscore}, which measures faithfulness through semantic alignment. 
For factuality evaluation, we employ FactScore~\cite{min-etal-2023-factscore}, which retrieves evidence from a knowledge base, and FacTool~\cite{chern2023factool}, which retrieves evidence through online searches. 
Faithfulness evaluation determines the presence of hallucinations, while factuality evaluation classifies them as factual or non-factual.

\paragraph{Knowledge Retrieval}
This approach involves only retrieving summary-relevant knowledge and integrating it directly into the context alongside the document, offering a straightforward method for incorporating external information. Specifically, we break the summary $S$ into individual claims $\{c_1, ..., c_l\}$ and retrieve knowledge evidence $E^{kb}$ related to the entities in each claim $c_i$.

\paragraph{Concurrent Retrieval}
This approach involves simultaneously retrieving evidence from the source document and the knowledge base for each claim, referred to as Concurrent Retrieval. 
Specifically, LLMs generate a query $q_i$ for each claim $c_i$. These queries are then used to retrieve the most relevant evidence from the evidence pool, which comprises the chunked source document $E^d$ and entity-centric knowledge evidence $E^{kb}$, as Eq.~\ref{eq:mr_doc} and Eq.~\ref{eq:mr_kb}.

\begin{align}
  e^d_i &= \operatorname*{argmax}_{e^d_j\in E^d} R(e^d_j \mid q_i),  \forall q_i\in\{q_1, ..., q_l\} \label{eq:mr_doc} \\
  e^{kb}_i &= \operatorname*{argmax}_{e^{kb}_j\in E^{kb}} R(e^{kb}_j \mid q_i),  \forall q_i\in\{q_1, ..., q_l\} \label{eq:mr_kb}
\end{align}

\paragraph{Reflection Retrieval}
The summary hallucination existence is primarily judged based on the source document.
Therefore, this approach adopts a two-stage retrieval evaluation. In the first stage, we keep Eq.~\ref{eq:mr_doc}, where evidence $e^d_i$ is retrieved from the source document to assess faithfulness. Based on this evaluation, the LLM reflects and identifies missing or incomplete information and generates a refined query $q^r_i$ to retrieve additional evidence from the knowledge base, as Eq.~\ref{eq:rr_query}. 
\begin{equation}
    q^r_i = P(y\mid [c_i, e^d_i], q_{reflect})  \label{eq:rr_query} 
\end{equation}

This sequential process ensures that hallucination is evaluated with a more comprehensive set of evidence, which we call it Reflection Retrieval.

\section{Experiment and Result}
\label{sec:experiment}
This section presents the experiment settings and offers a comprehensive analysis of LLM-based evaluators' performance in detecting mixed-context hallucinations. Through carefully designed experiments, we investigate our research questions while addressing critical challenges in hallucination evaluation using LLMs, while also proposing potential solutions.

\subsection{Datasets}
To evaluate the performance of LLMs on mixed-context hallucination evaluation, we use our newly introduced dataset, FHSumBench, and incorporate an additional dataset, M-XSum~\cite{maynez-etal-2020-faithfulness}, for comparable experiments. The difference between these two datasets is that FHSumBench is built based on knowledge fact injection, which can examine LLMs' ability to leverage internal parametric knowledge and external knowledge. 
Notably, the M-XSum dataset is constructed through multiple small masked language models generating summaries followed by human annotation, where non-factual hallucinations account for the vast majority (92\%), making a supplement of this category.

\subsection{Implementing Details}
\label{sec:experiment:implement}
\paragraph{LLM Selection}
We evaluate open-source models, including Llama~\cite{touvron2023Llamaopenefficientfoundation} and Qwen~\cite{yang2024Qwen2} families, across parameter sizes from 0.5B to 72B, alongside the closed-source GPT-4o, to examine how model capacity influences mixed-context hallucination evaluation. For retrieval-based judgment, we use Llama-Index\footnote{\url{https://github.com/run-llama/llama_index}} for retrieval embedding and retrieving the highest-scoring evidence at each step.
LLM inference is carried out using vLLM~\footnote{\url{https://github.com/vllm-project/vllm}} on 4 NVIDIA A100$^{80\texttt{GB}}$ GPUs.

\paragraph{Evaluation Settings}
For our retrieval-based approaches, each retrieved piece of evidence is assessed for its relevance (whether it is related to the claim) and supportiveness (whether it supports or contradicts the claim). The system then assigns a faithfulness score and a factuality score to each claim while maintaining a complete trace of the retrieval process. 
The individual claim scores are aggregated to evaluate the overall summary.
We report the overall macro precision, recall, and F1-score for all methods as well as the accuracy of each category. The full results can be seen in Appendix~\ref{apx: full results}, while a detailed analysis of the representative models is presented in \textsection\ref{Sec:Result}.

\begin{table*}[tbp]
\centering
\begin{adjustbox}{width=0.7\textwidth}
\begin{tabular}{cl|rrrrrr}
\Xhline{1.5px}
\multicolumn{2}{c|}{\multirow{2}{*}{\textbf{Methods}}} & \multicolumn{3}{c}{\textbf{FHSumBench}} & \multicolumn{3}{c}{\textbf{M-XSum}} \\ 
\multicolumn{2}{c|}{}  & \multicolumn{1}{c}{\textbf{P}} & \multicolumn{1}{c}{\textbf{R}} & \multicolumn{1}{c}{\textbf{F}} & \multicolumn{1}{c}{\textbf{P}} & \multicolumn{1}{c}{\textbf{R}} & \multicolumn{1}{c}{\textbf{F}} \\ \Xhline{1.5px}
\multirow{4}{*}{Hybrid} & FS*WC & 0.3914 & 0.3073 & 0.3443 & 0.3522 & 0.3458 & 0.3490 \\
 & FS*AS & 0.3870 & 0.3366 & 0.3601 & 0.3470 & 0.2690 & 0.3030 \\
 & FT*WC & 0.3508 & 0.3604 & 0.3555 & 0.4037 & 0.3715 & 0.3869 \\
 & FT*AS & 0.3705 & 0.4034 & 0.3863 & 0.4069 & 0.2959 & 0.3426 \\ \hline
\multirow{3}{*}{KR} & Llama3-8B & 0.3548 & 0.2929 & 0.3209 & 0.3575 & 0.2298 & 0.2798 \\
 & Qwen2.5-14B & 0.4628 & 0.4740 & 0.4683 & 0.4035 & 0.4716 & 0.4349 \\
 & GPT-4o & 0.4669 & 0.4763 & 0.4715 & 0.3801 & 0.4009 & 0.3902 \\ \hline
\multirow{3}{*}{CR} & Llama3-8B & 0.4461 & 0.4420 & 0.4441 & \underline{0.4794} & 0.4470 & 0.4626 \\
 & Qwen2.5-14B & \textbf{0.5395} & 0.4167 & 0.4702 & 0.4579 & 0.4590 & 0.4584\\
 & GPT-4o & 0.5026 & \underline{0.4878} & \underline{0.4951} & 0.4214 & 0.4367 & 0.4289 \\ \hline
\multirow{3}{*}{RR} & Llama3-8B & 0.4664 & 0.4615 & 0.4640 & \textbf{0.4869} & \underline{0.4752} & \textbf{0.4810} \\
 & Qwen2.5-14B & \underline{0.5325} & 0.4148 & 0.4663 & 0.4766 & \textbf{0.4787} & \underline{0.4776} \\
 & GPT-4o & 0.5135 & \textbf{0.4891} & \textbf{0.5010} & 0.4413 & 0.4613 & 0.4511 \\ \Xhline{1.5px}
\end{tabular}
\end{adjustbox}
\caption{The results using retrieval-based LLM evaluators on FHSumBench and M-XSum datasets. We report the precision (P), recall (R) and F1-score (F) in this table. The best and second-best results are highlighted in \textbf{bold} and \underline{underlined}, respectively. Hybrid scores include the combination of FactScore (FS), FacTool (FT), WeCheck (WC) and AlignScore (AS). "KR" denotes knowledge retrieval, "CR" denotes concurrent retrieval, and "RR" denotes reflection retrieval.}
\label{tab:result_rq2}
\end{table*}

\subsection{Results and Analysis}
\label{Sec:Result}

\paragraph{RQ1: How do the LLMs perform in hallucination evaluation?}

Table~\ref{tab:result_rq1} presents the evaluation results on the FHSumBench and M-XSum datasets, comparing baselines and LLM direct generation methods. 
The random score on both datasets presents a lower bound for the evaluation.
The performance of vanilla judge varies significantly across models. GPT-4o consistently outperforms the other two LLMs. However, even in the best case, vanilla LLM judges exhibit moderate F1-scores, indicating room for improvement.

By incorporating different prompting strategies, most models achieve improvement over the vanilla judge baseline. Specifically, under the +ICL setting, Llama3-8B achieves an F1-score improvement of 12.3\% and 33.5\%, respectively, highlighting that few-shot examples offer greater advantages to smaller LLMs by serving as a crucial guide to task comprehension.
CoT leads to noticeable gains, particularly for Qwen2.5-14B, which achieves the highest F1-score (0.4733) on FHSumBench. 
However, on M-XSum, CoT slightly underperforms for larger models due to knowledge bias and hallucinations in reasoning (see \textsection\ref{sec:case} for details).

\paragraph{RQ2: How can LLMs better leverage external knowledge?}

Overall, concurrent retrieval and reflective retrieval methods demonstrate superior performance compared to the other two approaches.
However, knowledge retrieval, which simply integrates all entity knowledge in the summary, still shows improvement compared to direct generation evaluation. 
The advantage of retrieval-based methods likely comes from providing explicit external knowledge, thereby reducing the LLM’s reliance on potentially incomplete or incorrect intrinsic knowledge.
For mixed-context evidence retrieval, allowing the LLM to perform step-by-step retrieval by identifying missing information during the retrieval process proves to be more effective.
FactScore utilizes Wikipedia database for retrieving knowledge evidence. However, its retrieval efficiency is relatively low, with only 55.2\% of entities in FHSumBench and 45\% in M-XSum successfully retrieving relevant knowledge. This significantly influences the accuracy of the evaluation. To address this, we employ GPT-4o to generate knowledge for entities and substantially improve evaluation performance (detailed analysis available in Appendix~\ref{apx:gpt_knowledge}).
FacTool relies on online search to retrieve evidence, resulting in more noisy information. Consequently, the quality of evidence interpretation significantly impacts the final evaluation results.

\begin{table}[t]
\centering
\begin{adjustbox}{width=0.48\textwidth}
\begin{tabular}{llc|cccc}
\hline
 &  &   & \textbf{P} & \textbf{R} & \textbf{F} \\ \hline
 
\multicolumn{2}{l}{Reflection Retrieve}  &  & \cellcolor{gray!20} 0.4664  & \cellcolor{gray!20}0.4615 & \cellcolor{gray!20}0.4640 \\ \hline
   &\multirow{1}{*}{KB} & Wiki  & 0.4200 & 0.3611 & 0.3129 \\
 \hline
   &Reflection & loop & 0.4532 & 0.4325 & 0.4156 \\ \hline
   &Top-k & k=3 & 0.4631 & 0.4613 & 0.4599 \\ \hline
\end{tabular}
\end{adjustbox}
\caption{Ablation study of the RR method on FHSumBench. "Wiki" indicates using Wikidata to retrieve knowledge evidence. "Loop" in reflection refers to iterative evidence retrieval until the LLM can provide a definitive True or False judgment. "Top-k" specifies the retrieval of $k$ evidence chunks from the evidence pool.}
\label{tab:ablation}
\end{table}

Compared to direct generation evaluation, Llama3-8B has the highest improvement over all the methods on both datasets, which indicates that smaller models can benefit more from evidence retrieval.
We also conducted an ablation study on the experimental setup of reflection retrieval (see Table~\ref{tab:ablation}). The results indicate that using Wikidata~\cite{10.1145/2629489} as the knowledge base faces the same challenge as FactScore, where the evaluation results are heavily influenced by the entity matching rate. 
Additionally, we found that in this task, since both the document and knowledge are of moderate length, retrieving the top $k=3$ results yields similar outcomes as retrieving only the top $k=1$. Furthermore, we experimented with allowing the LLM to iteratively reflect until it retrieves sufficient evidence to draw a conclusion of either "support" or "contradict". However, we observed that during this process, the LLM exhibited a tendency to overcorrect its judgments.

\begin{figure}[tbp]
    \centering
    \subfigure{
        \includegraphics[width=0.47\textwidth]{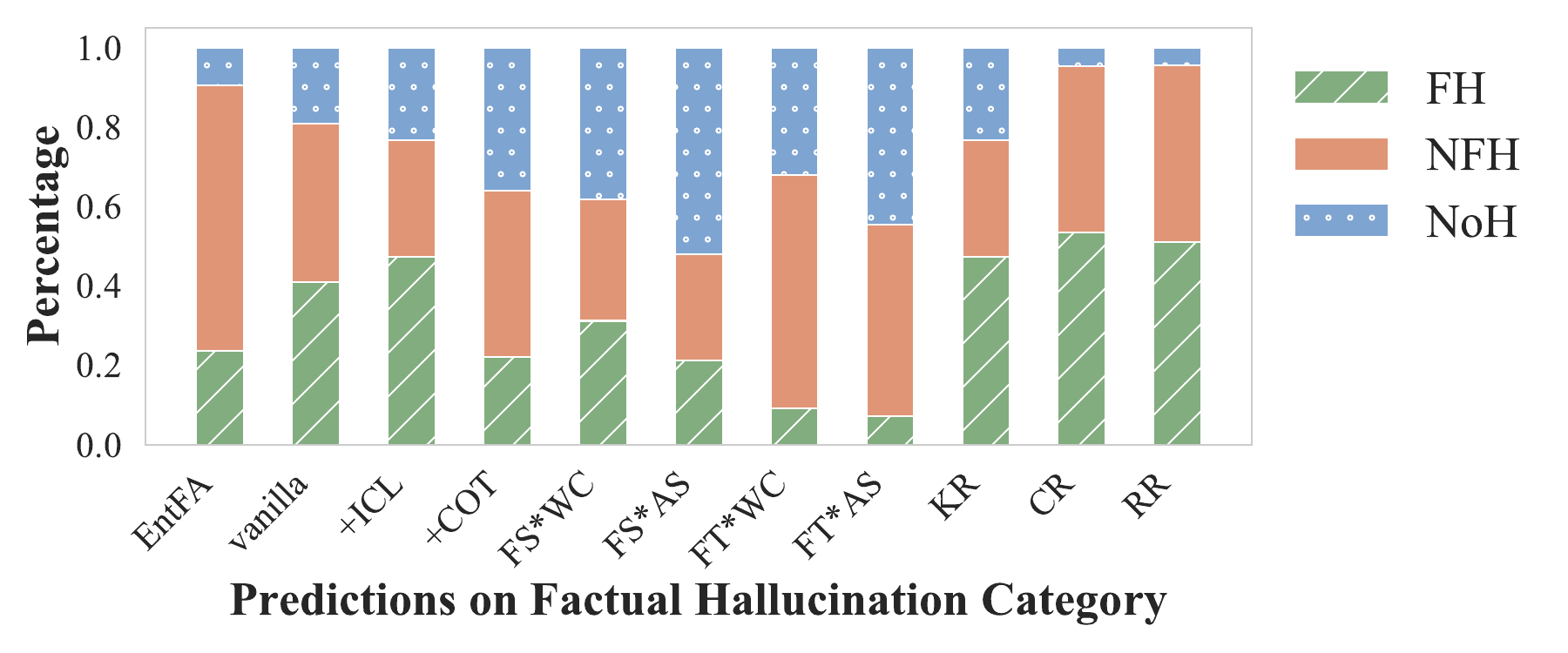} 
        \label{fig:stack_bar1}}
    \subfigure{
        \includegraphics[width=0.47\textwidth]{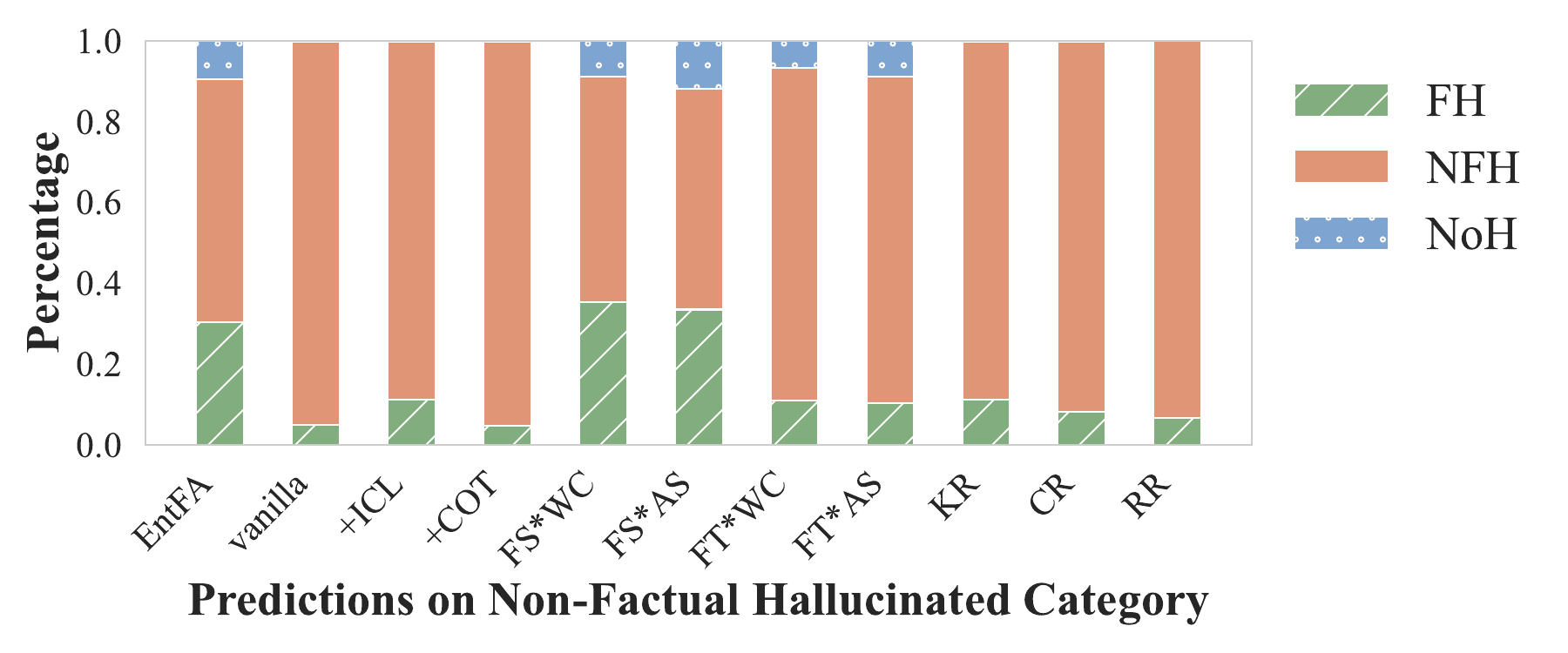} 
        \label{fig:stack_bar2}}
    \subfigure{
        \includegraphics[width=0.47\textwidth]{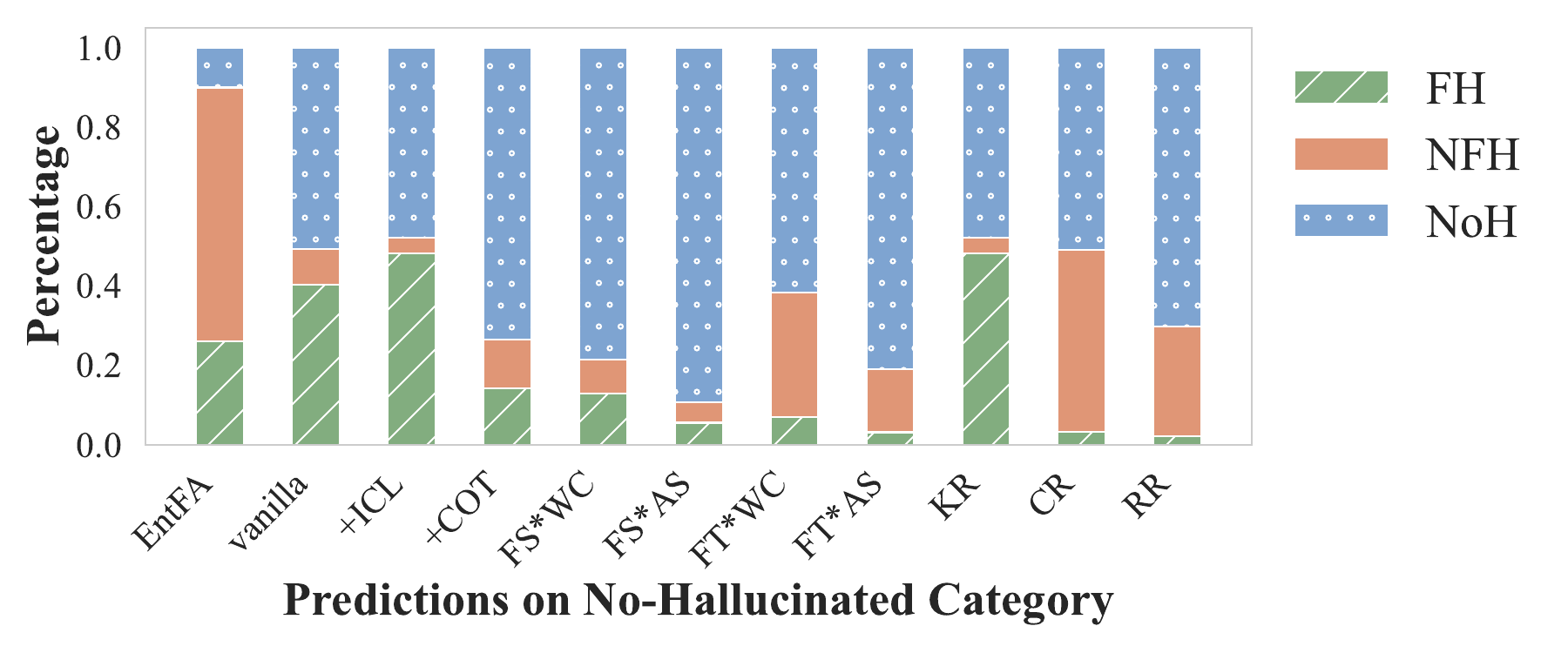}
        \label{fig:stack_bar3}}
    \caption{Percentages of the GPT-4o predictions on each category of FHSumBench.}
    \label{fig:stack_bar}
\end{figure}

\paragraph{RQ3: Which category dominates the results?}

We selected GPT-4o to analyze model predictions across three data categories (details of other models available in Appendix~\ref{apx: full results}).
As shown in Figure~\ref{fig:stack_bar}, we report the prediction percentage for each category. 
Our analysis reveals a clear pattern in hallucination detection performance. 
For factual hallucinations, retrieval-based approaches demonstrate the highest accuracy, indicating that retrieving different contextual evidence helps detect mixed-context hallucinations.
However, compared to a vanilla judge, CoT reasoning tends to classify more factual hallucinations as "no hallucination," suggesting that, in this case, LLMs are likely to deviate from predefined evaluation criteria during the reasoning process.

Additionally, most methods exhibit high sensitivity to non-factual hallucinations, indicating that unfaithful text is more likely to be classified as non-factual rather than factual. This behavior may be attributed to less conflict between LLMs' intrinsic knowledge and external knowledge when evaluating non-factual hallucinations.

While CoT reasoning significantly improves the accuracy of no-hallucination predictions, simpler approaches such as ICL show only marginal improvements over the vanilla judge. 
Furthermore, although the reflection retrieval judge demonstrates superior performance on non-hallucinated data compared to the concurrent retrieval judge, our comprehensive analysis indicates a notable trade-off: the improved accuracy in detecting factual hallucinations comes at the cost of decreased performance in the other two categories.
The model size also plays a crucial role in evaluation stability, while larger LLMs show a more stable performance across the methods on each category (according to Figure~\ref{fig:stack_bar_8},\ref{fig:stack_bar_14},\ref{fig:stack_bar_32} in Appendix~\ref{apx: full results}).

\begin{figure}[t]
  \centering
  \includegraphics[width=0.47\textwidth]{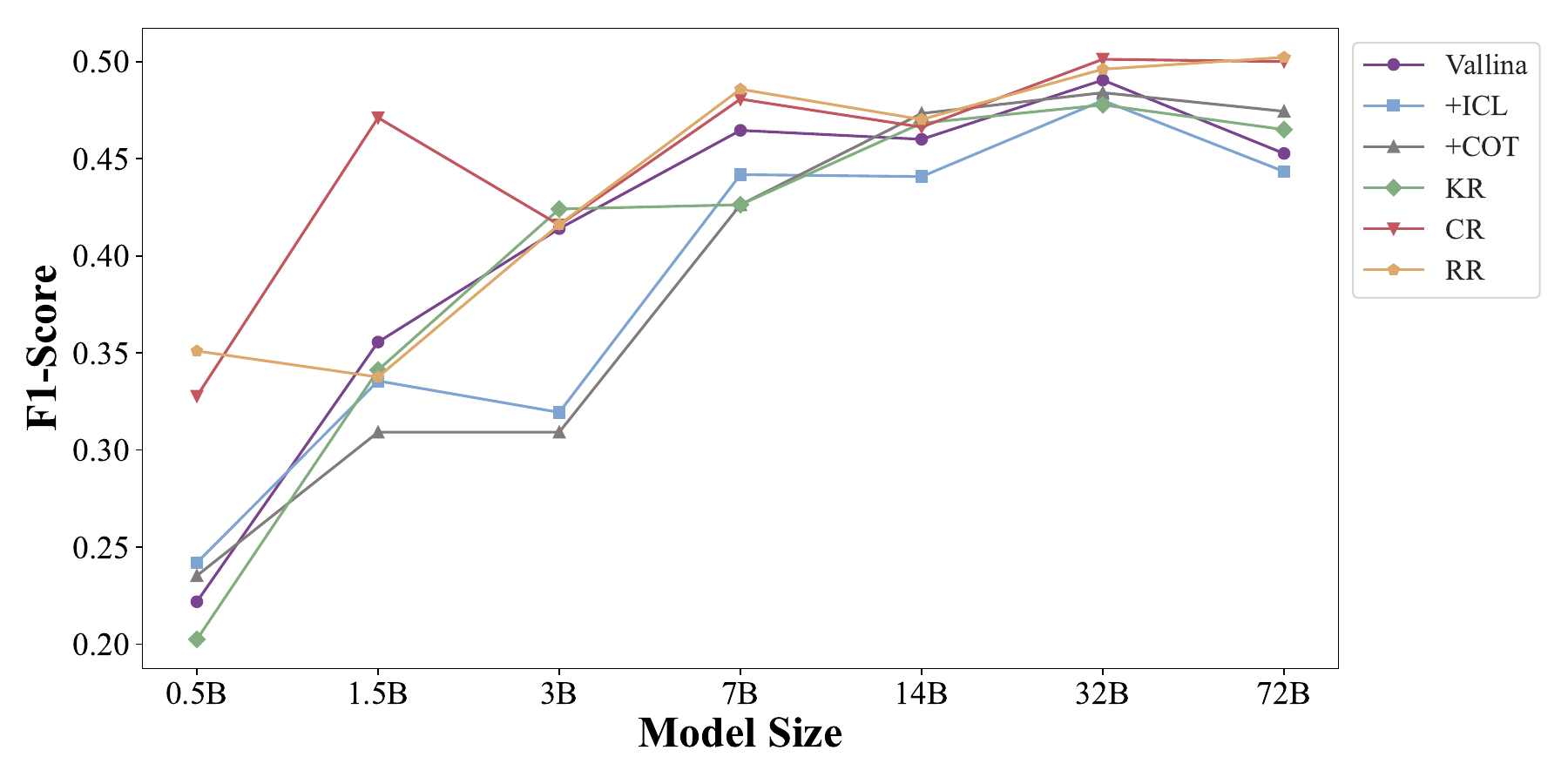}
  \caption{Model performance on F1-score with the increasing of model size in Qwen2.5 families.}
  \label{fig:scale}
\end{figure}

\paragraph{RQ4: Can scaling resolve the problem?}
Figure~\ref{fig:scale} demonstrates the performance of the best-performing reflection retrieve approach across the Qwen2.5 family models of varying sizes. 
Performance improves as the model size increases, with larger models (e.g., Qwen2.5-32B) achieving significantly higher overall accuracy, particularly in the factual hallucination category (see Table~\ref{tab:full_res_scale} in Appendix~\ref{apx: full results}), indicating that reflection with LLMs can help better identify and utilize the evidence in different context.
However, the results from both Qwen2.5-72B and GPT-4o demonstrate that increased model size does not necessarily correlate with improved performance. We will further investigate the reason in \textsection\ref{sec:case}.

\section{Case Study}
\label{sec:case}

We choose the representative larger LLM, GPT-4o, as the case study model to investigate the reasons for the errors in hallucination detection (examples available in Appendix~\ref{apx:case_study}).

\paragraph{Intrinsic knowledge bias of LLM judge}
Since the CoT method exhibits the lowest score in detecting factual hallucinations using GPT-4o among direction generation LLM methods, we selected it to investigate the causes of error cases. We randomly selected and analyzed 30 error cases from the CoT outputs of GPT-4o. 
Among these cases, 63.3\% were misclassified as `no hallucination' due to the LLM either relying on its intrinsic knowledge for reasoning or inherently assuming the injected information to be faithful. Additionally, 30\% of the samples were mistakenly categorized as `non-factual hallucination' due to insufficient evidence being provided.
The remaining 6.7\% of errors resulted from issues in the reasoning process. These cases include instances where unfaithful content was correctly flagged as hallucination during the CoT process but was later misinterpreted as faithful in the final judgment. 
Conversely, some cases involved injected factual knowledge being accurately recognized during reasoning but ultimately misjudged as false in the factuality assessment.
This indicates that, despite well-defined evaluation criteria, the model still struggles to some extent to distinguish between faithfulness and factuality.

\paragraph{The influence of retrieval}
Since the intrinsic knowledge of the LLM significantly influences its decision-making, retrieval-based experiments demonstrate that it can partially rectify erroneous classifications. To further investigate this, we examined the results of the CR and RR methods on GPT-4o using the aforementioned samples.
Overall, evidence retrieval helps mitigate intrinsic knowledge bias. Among the analyzed error cases, the CR method successfully corrects 47.3\% of samples misclassified as "no hallucination", while the RR method achieves a correction rate of 57.9\%.
However, the effectiveness of this approach still depends on the appropriateness of the knowledge base and the accuracy of retrieval. Although the RR method improves performance, there remains room for further enhancement.

\section{Conclusion and Future Direction}

In this paper, we systematically evaluate the ability of LLMs to assess hallucinations in mixed-context scenarios within summarization. The evaluation covers both open-source and closed-source models across various scales. 
Specifically, the distinction between factual and non-factual hallucinations is analyzed across different models and methods. 
Our findings reveal that LLMs still struggle to differentiate between faithfulness and factuality, highlighting the benefit of explicitly incorporating external knowledge to aid judgment. While larger models generally exhibit better performance, the trend suggests the existence of a performance plateau. Drawing insights from our experimental findings, several directions deserve further investigation:

\paragraph{Reducing Intrinsic Knowledge Bias}
Our experiments reveal that intrinsic knowledge bias significantly impacts LLM evaluation results, particularly for larger LLMs. While we have conducted preliminary investigations, deeper exploration is needed to understand fundamental aspects such as training data updates, internal information flow, and mechanisms for integrating internal and external knowledge.

\paragraph{Effective Knowledge Integration}
Mixed-context hallucination evaluation requires the fusion of multiple sources of information for decision-making. Our experiments demonstrate that specific information integration methods enable smaller LLMs to achieve performance comparable to larger LLMs. This highlights the urgent need for research into more effective information fusion approaches.

\paragraph{More Flexible Scenarios}
While this work primarily investigates mixed-context hallucination evaluation through the lens of summarization tasks, real-world scenarios often present greater complexity and flexibility. These include extended dialogues and more intricate task workflows that extend beyond knowledge-based hallucination. Further research is needed to address these diverse applications.

\section*{Limitation}

\paragraph{Knowledge-Based Hallucination Construction}
While our approach of leveraging entity knowledge from Wikipedia to construct factual and non-factual hallucinations proved effective, certain limitations remain. The current work does not address discourse-level hallucinations or event-based hallucinations, which still paves the way for future research. 
Additionally, since the contexts in news article summarization are factually correct, we did not further explore cases where a summary is faithful but not factual. Despite these constraints, our approach maintains broad practical applicability.

\paragraph{Benchmark Hacking}
Our benchmark data primarily incorporates hallucinations through appositive constructions. This structural consistency suggests that targeted training could potentially yield overfitting models that excel at evaluating such patterns. However, our primary objective is to assess LLMs' capability to differentiate mixed-context hallucinations, rather than to simulate model-generated data.

\paragraph{LLM-as-a-Judge Setting}
Although we extensively explored various common methods for employing LLMs as judges, our investigation was limited to text-based approaches. The potential of utilizing LLM prediction uncertainty for hallucination evaluation remains unexplored and presents a promising direction for future research.

\section*{Ethics Statement}
Our study evaluates LLMs assessment of mixed-context hallucination using summarization datasets. We conduct experiments on both our constructed dataset and widely used datasets. Specifically, our dataset is derived from CNN/DailyMail and XSum, which contain news articles and summaries primarily in English. As such, our study focuses exclusively on English-language scenarios, which may limit the applicability to other languages and cultural contexts.
Future research could incorporate more linguistically and culturally diverse datasets, and improve representation across different racial, ethnic, and cultural backgrounds, ultimately contributing to more equitable and globally relevant NLP models.

\section*{Acknowledgments}
SQ is funded by a PhD scholarship provided by CSC. 
The authors acknowledge Wei Liu for the advice on formatting this paper,
and the use of King’s Computational Research, Engineering and Technology Environment (CREATE) at King’s College London.

\bibliography{acl_latex}

\clearpage
\appendix

\label{sec:appendix}

\section{Supplementary Experiment Details}

\subsection{Dataset Details}
\label{apx:dataset}
Table~\ref{tab:dataset_info} shows the data size, text length, and injected-fact length of each category in FHSumBench.
Here we also present the meta-information of FHSumBench and previous related datasets on hallucinations of summarization evaluation. The relationship between different datasets is shown in Table~\ref{tab:dataset_compare}.

\begin{table}[htbp]
\centering
\begin{adjustbox}{width=0.45\textwidth}
\begin{tabular}{ccccc}
\Xhline{1.5px}
 &
  \begin{tabular}[c]{@{}c@{}}\textbf{Factual}\\\textbf{Hallu}\end{tabular} &
  \begin{tabular}[c]{@{}c@{}}\textbf{Non-Factual}\\\textbf{Hallu}\end{tabular} &
  \begin{tabular}[c]{@{}c@{}}\textbf{No-}\\\textbf{Hallu}\end{tabular} &
  \textbf{Overall} \\ \Xhline{1.5px}
Data Size& 501 & 410 & 425 & 1336 \\ \hline
  \#Doc  & 433 & 448 & 471 & 449  \\ \hline
\#Sum   & 29  & 27  & 20  & 26   \\ \hline
\#Inject & 6   & 4   & -   & 5    \\ \Xhline{1.5px}
\end{tabular}
\end{adjustbox}
\caption{Meta information of FHSumBench, including the average word number per document (\#Doc), summary (\#Sum) and the injected fact (\#Inject).}
\label{tab:dataset_info}
\end{table}

Constructed from XEnt and FactCollect by fact injection, FHSumBench comprises news articles from both XSum and CNN/DM, with summaries in CNN/DM being longer than those in XSum.

\begin{table*}[htbp]
\centering
\begin{adjustbox}{width=\textwidth}

\begin{tabular}{ccccc}
\Xhline{1.5px}
\multicolumn{1}{l}{} &
  \textbf{Label Categories} &
  \textbf{Test Set Size} &
  \textbf{Data Source} &
  \textbf{Annotate Granularity} \\\Xhline{1.5px}
FHSumBench & \begin{tabular}[c]{@{}c@{}c@{}}factual hallucination, \\ non-factual hallucination, \\no hallucination\end{tabular}      & 1336 & XEnt, FactCollect & summary \\ \hline
XEnt~\cite{cao-etal-2022-hallucinated}    & \begin{tabular}[c]{@{}c@{}c@{}}factual hallucination, \\ non-factual hallucination, \\no hallucination\end{tabular}      & 240  & M-XSum             & entity  \\ \hline
FactCollect~\cite{ribeiro-etal-2022-factgraph} & hallucination, no hallucination & 502 &
  \begin{tabular}[c]{@{}c@{}}FactCC, M-XSum, \\ QAGS, Frank\end{tabular} & summary \\ \hline
M-XSum~\cite{maynez-etal-2020-faithfulness}  &
  \begin{tabular}[c]{@{}c@{}}intrinsic/extrinsic hallucination, \\ factual/unfactual\end{tabular} & 912 & XSum & span, summary \\ \hline
Frank~\cite{pagnoni-etal-2021-understanding}   & \begin{tabular}[c]{@{}c@{}}semantic frame errors,  discourse errors,  \\ content verifiability errors\end{tabular} & 350  & CNN/DM, XSum      & span    \\ \hline
QAGS~\cite{wang-etal-2020-asking} & hallucination, no hallucination & 474 & CNN/DM, XSum & summary \\ \hline
FactCC~\cite{kryscinski-etal-2020-evaluating} & hallucination, no hallucination & 503 & CNN/DM & claim \\ \Xhline{1.5px}
\end{tabular}
\end{adjustbox}
\caption{The meta information and relationship between different datasets on faithfulness and factuality of summarization evaluation.}
\label{tab:dataset_compare}
\end{table*}

\subsection{Previous Work Settings}
\label{apx:baselines}
Here we present the detailed descriptions and settings of the baselines in the experiments.

\paragraph{EntFA}
The EntFA~\cite{cao-etal-2022-hallucinated} method aims to distinguish between factual hallucination and non-factual hallucination in abstractive summarization by evaluating both faithfulness and factuality as well. It performs evaluation at the entity level, utilizing both prior and posterior features of entities, which are extracted from a conditional masked language model. These two features are then used to train two separate KNN models to assess faithfulness and factuality individually. We label entities within the hallucination span as "hallucinated" and those outside the span as "non-hallucinated". The models then classify the faithfulness and factuality for each entity, which are used to determine the final categorization.

\paragraph{WeCheck}
WeCheck~\cite{wu-etal-2023-wecheck} is a consistency checker which verifies the alignment between generated summaries and their source documents. It employs weakly supervised learning to train a model that detects inconsistencies by leveraging large-scale noisy datasets and learning from multiple NLI models. The workflow includes extracting key facts from the source, comparing them with the summary, and scoring the consistency. The model is trained based on DeBERTaV3~\cite{hedebertav3} to evaluate the faithfulness of the summaries.

\paragraph{AlignScore}
AlignScore~\cite{zha2023alignscore} is a metric for evaluating the factual consistency of summaries by aligning information between the summary and the source document. It works by leveraging pretrained language models to compute semantic alignments and generate a consistency score that reflects how well the summary captures the source's key facts. The pretrained model used in AlignScore is RoBERTa~\cite{liu2019roberta} models (125M and 355M), which is fine-tuned on the task of faithfulness evaluation. 

\paragraph{FactScore}
FactScore~\cite{min-etal-2023-factscore} is an evaluation framework designed to assess the factual accuracy of text by assigning a score based on atomic fact veracity. It operates by breaking down content into claims, comparing them against knowledge bases. It uses a GPT-series model to extract claims and a Wikipedia knowledge base to retrieve evidence. We use GPT-4o to keep the consistency with our LLM-based experiments. 
Note that in FactScore, defining a topic for the input text is necessary. We identify entities in the summary as topics and then retrieve the corresponding evidence passages accordingly.

\paragraph{FacTool}
FacTool~\cite{chern2023factool} is a tool designed for fact-checking to verify the factuality of statements by cross-referencing them with online search results. It supports multiple subtasks, among which we select the most relevant one, namely knowledge-based QA. FacTool operates by extracting claims, utilizing the Google Search API~\footnote{https://serper.dev/} to search the top pages and retrieve the most relevant search snippets, and then providing a confidence score for each validation. We use GPT-4o to extract claims and make judgments in our experiments to maintain the consistency of the setting.

\subsection{LLM-based methods}

Following FacTool and FactScore, claims for retrieval-based methods are generated by GPT-4o, with coreference resolution applied. The final classification is determined based on both the faithfulness score and the factuality score, where factual hallucination means unfaithful but factual, non-factual hallucination means unfaithful and unfactual, while no hallucination means both faithful and factual.
For direct generation, evaluators directly provide these scores. As mentioned in \textsection\ref{sec:experiment:implement}, retrieval-based methods aggregate results from individual claims. If any claim contains a non-factual hallucination, the summary is classified as a non-factual hallucination. Similarly, if any claim contains a factual hallucination, the summary is classified as a factual hallucination. Otherwise, if all claims remain faithful to the source document, it is classified as no hallucination.

\begin{table*}[htbp]
\centering
\begin{adjustbox}{width=0.85\textwidth}
\begin{tabular}{clcccccc}
\Xhline{1.5px}
\multicolumn{2}{c}{\multirow{2}{*}{\textbf{Methods}}} & \multicolumn{3}{c}{\textbf{FHSumBench}} & \multicolumn{3}{c}{\textbf{M-XSum}} \\ 
\multicolumn{2}{c}{} & \multicolumn{1}{l}{\textbf{FH-Acc}} & \multicolumn{1}{l}{\textbf{NFH-Acc}} & \multicolumn{1}{l}{\textbf{NoH-Acc}} & \multicolumn{1}{l}{\textbf{FH-Acc}} & \multicolumn{1}{l}{\textbf{NFH-Acc}} & \multicolumn{1}{l}{\textbf{NoH-Acc}} \\ \hline
\multicolumn{2}{l}{Random} & 0.3167 & 0.3374 & 0.3012 & 0.3190 & 0.3171 & - \\
\multicolumn{2}{l}{EntFA} & 0.2362 & 0.6 & 0.0996 & 0.3214 & 0.7252 & - \\ \hline
\multirow{3}{*}{Vallina Judge} & LLama-8B & 0.1198 & 0.2756 & \underline{0.7953} & 0.2154 & 0.3588 & - \\
 & Qwen-14b & 0.4092 & 0.7902 & 0.6612 & 0.5429 & 0.7280 & - \\
 & GPT-4o & 0.4052 & \underline{0.9488} & 0.5059 & 0.5000 & 0.8397 & - \\ \hline
\multicolumn{1}{r}{\multirow{3}{*}{+ICL}} & LLama-8B & 0.2794 & 0.4463 & 0.7153 & 0.3000 & 0.5982 & - \\
\multicolumn{1}{r}{} & Qwen-14b & 0.4072 & 0.9000 & 0.4871 & 0.4714 & 0.8325 & - \\
\multicolumn{1}{r}{} & GPT-4o & 0.4711 & 0.8854 & 0.4776 & \textbf{0.6857} & 0.7031 & - \\ \hline
\multicolumn{1}{r}{\multirow{3}{*}{+COT}} & LLama-8B & 0.1956 & 0.3512 & 0.6706 & 0.4211 & 0.4852 & - \\
\multicolumn{1}{r}{} & Qwen-14b & 0.2595 & 0.8756 & 0.7647 & 0.3478 & 0.8245 & - \\
\multicolumn{1}{r}{} & GPT-4o & 0.2176 & 0.9439 & 0.7341 & 0.2571 & 0.8728 & - \\ \hline
\multirow{4}{*}{Hybrid} & FS*WC & 0.2535 & 0.3829 & 0.5929 & 0.2429 & 0.7945 & - \\
 & FS*AS & 0.1756 & 0.3707 & \textbf{0.8000} & 0.2000 & 0.6069 & - \\
 & FT*WC & 0.0878 & 0.7610 & 0.5929 & 0.2000 & \textbf{0.9145} & - \\
 & FT*AS & 0.0699 & 0.7439 & \textbf{0.8000} & 0.1857 & 0.7019 & - \\ \hline
\multirow{3}{*}{KR} & LLama-8B & 0.0878 & 0.3122 & 0.7718 & 0.1667 & 0.5227 & - \\
\multicolumn{1}{r}{} & Qwen-14b & 0.3174 & 0.9244 & 0.6541 & \underline{0.5857} & 0.8290 & - \\
\multicolumn{1}{r}{} & GPT-4o & 0.2495 & \textbf{0.9732} & 0.6824 & 0.3286 & 0.8741 & - \\ \hline
\multirow{3}{*}{CR} & LLama-8B & \textbf{0.6128} & 0.6756 & 0.2188 & 0.2273 & 0.6667 & - \\
 & Qwen-14b & 0.4970 & 0.7927 & 0.3694 & 0.0588 & 0.8591 & - \\
 & GPT-4o & 0.4770 & 0.8293 & 0.4776 & 0.0000 & 0.8733 & - \\ \hline
\multirow{3}{*}{RR} & LLama-8B & \underline{0.5908} & 0.6878 & 0.1859 & 0.2069 & 0.7435 & - \\
 & Qwen-14b & 0.4711 & 0.8146 & 0.3812 & 0.0789 & 0.8784 & - \\
 & GPT-4o & 0.4393 & 0.8290 & 0.6538 & 0.0238 & \underline{0.8988} & - \\ \Xhline{1.5px}
\end{tabular}
\end{adjustbox}
\caption{Full results for each category in FHSumBench and M-XSum, including factual hallucination accuracy (FH-Acc), non-factual hallucination accuracy (NFH-Acc), and no-hallucination accuracy (NoH-Acc).}
\label{tab:full_categories}

\end{table*}

\section{Full Results}
\label{apx: full results}
\begin{figure}[]
    \centering
    \subfigure{
        \includegraphics[width=0.47\textwidth]{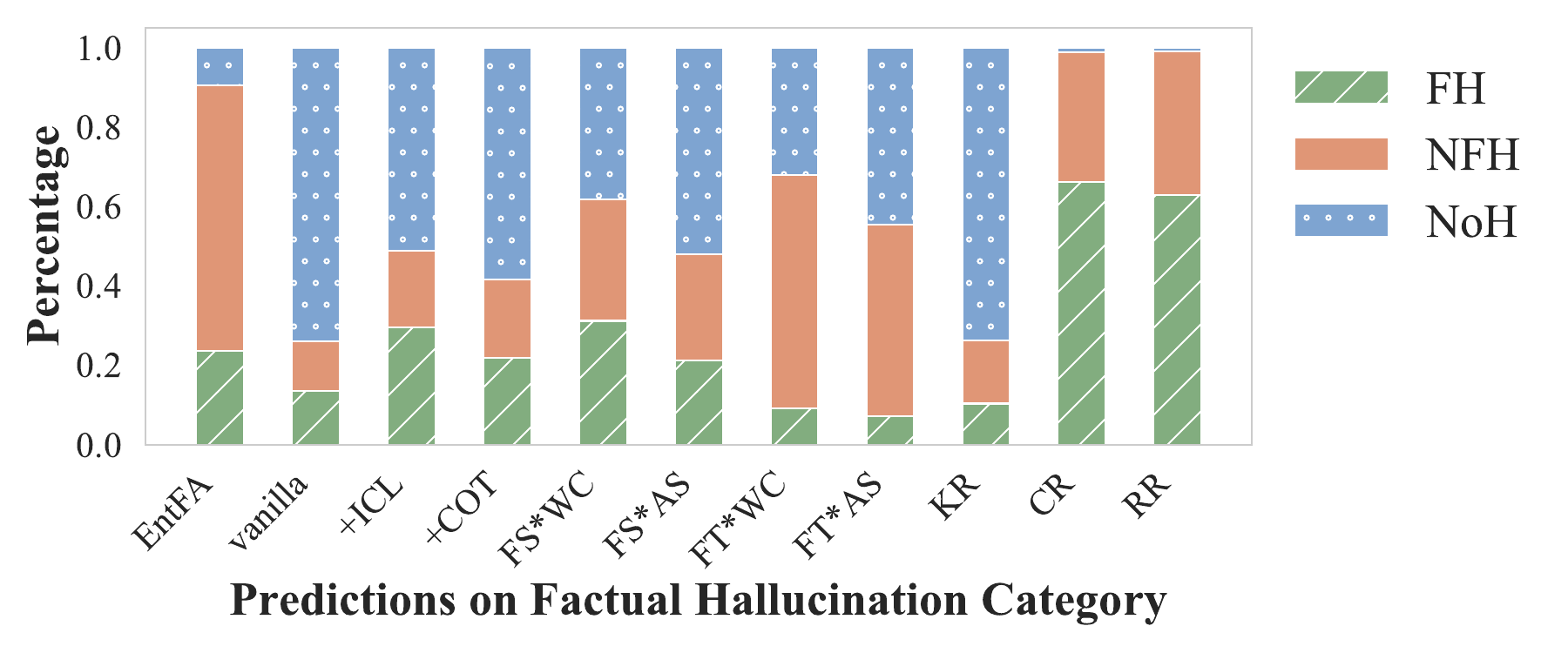} 
        \label{fig:stack_bar_8_1}}
    \subfigure{
        \includegraphics[width=0.47\textwidth]{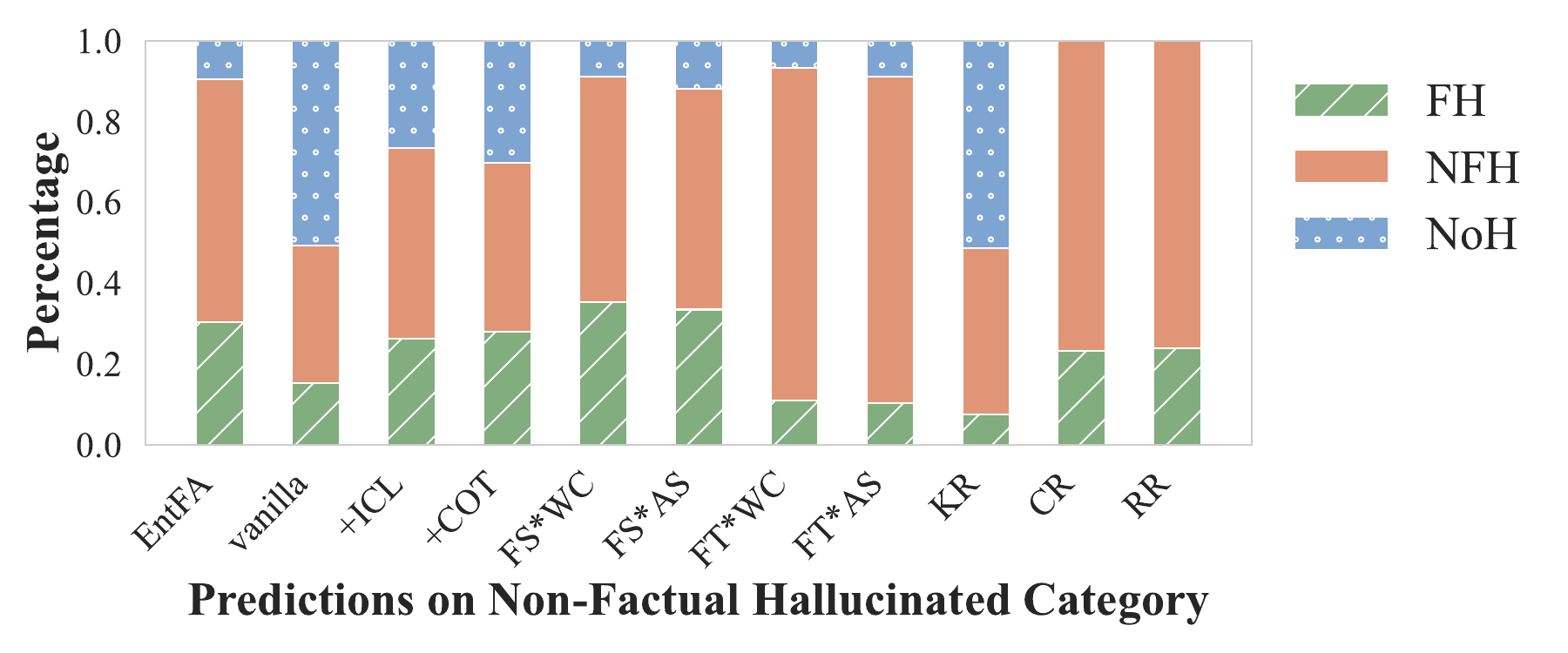} 
        \label{fig:stack_bar_8_2}}
    \subfigure{
        \includegraphics[width=0.47\textwidth]{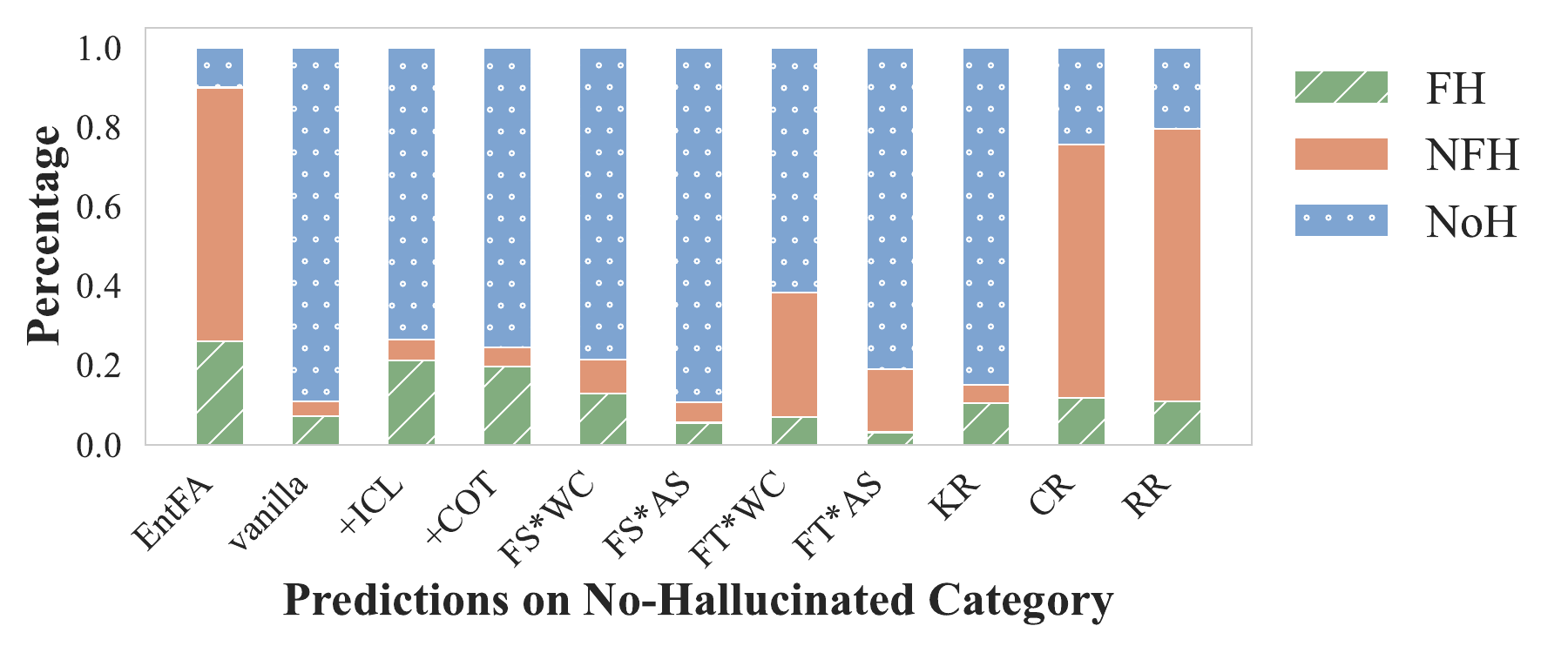}
        \label{fig:stack_bar_8_3}}
    \caption{Predictions of Llama3-8B on categories of FHSumBench.} 
    \label{fig:stack_bar_8}
\end{figure}

\begin{figure}[h]
    \centering
    \subfigure{
        \includegraphics[width=0.47\textwidth]{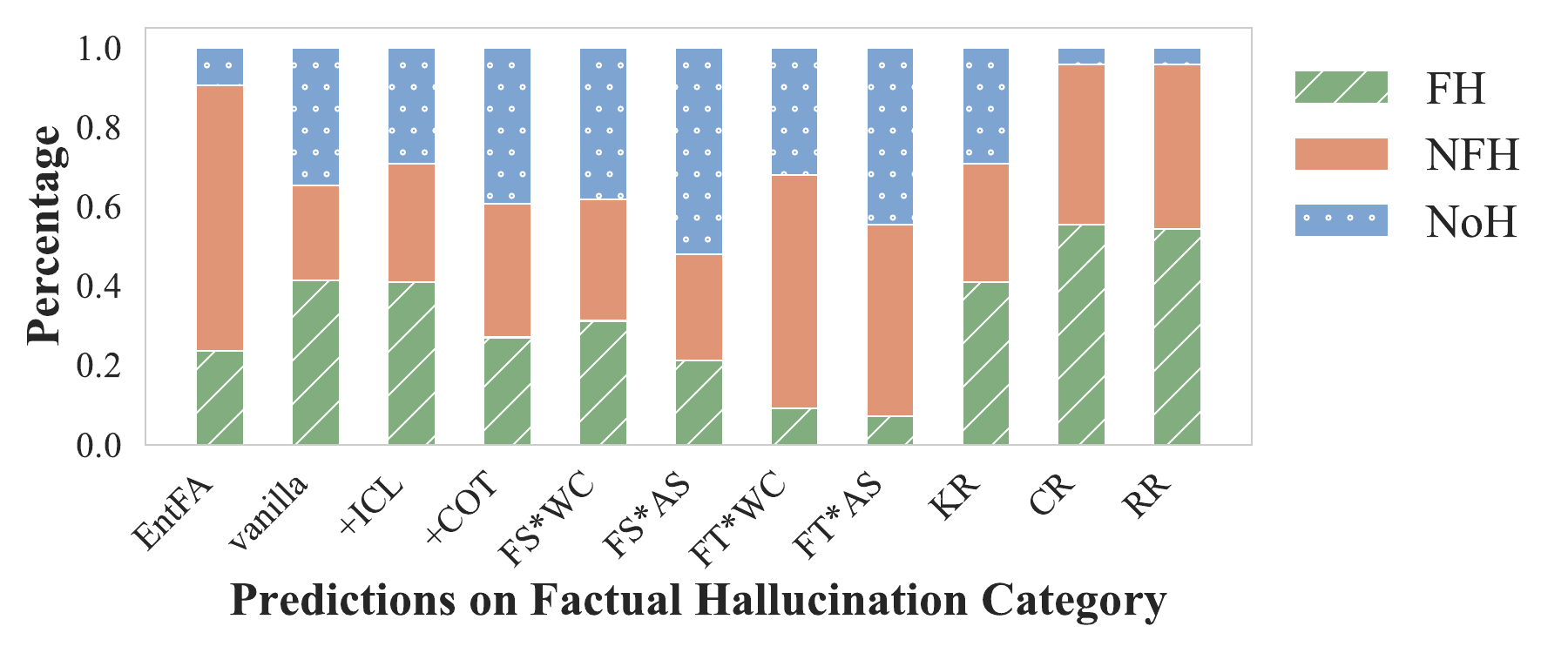} 
        \label{fig:stack_bar_14_1}}
    \subfigure{
        \includegraphics[width=0.47\textwidth]{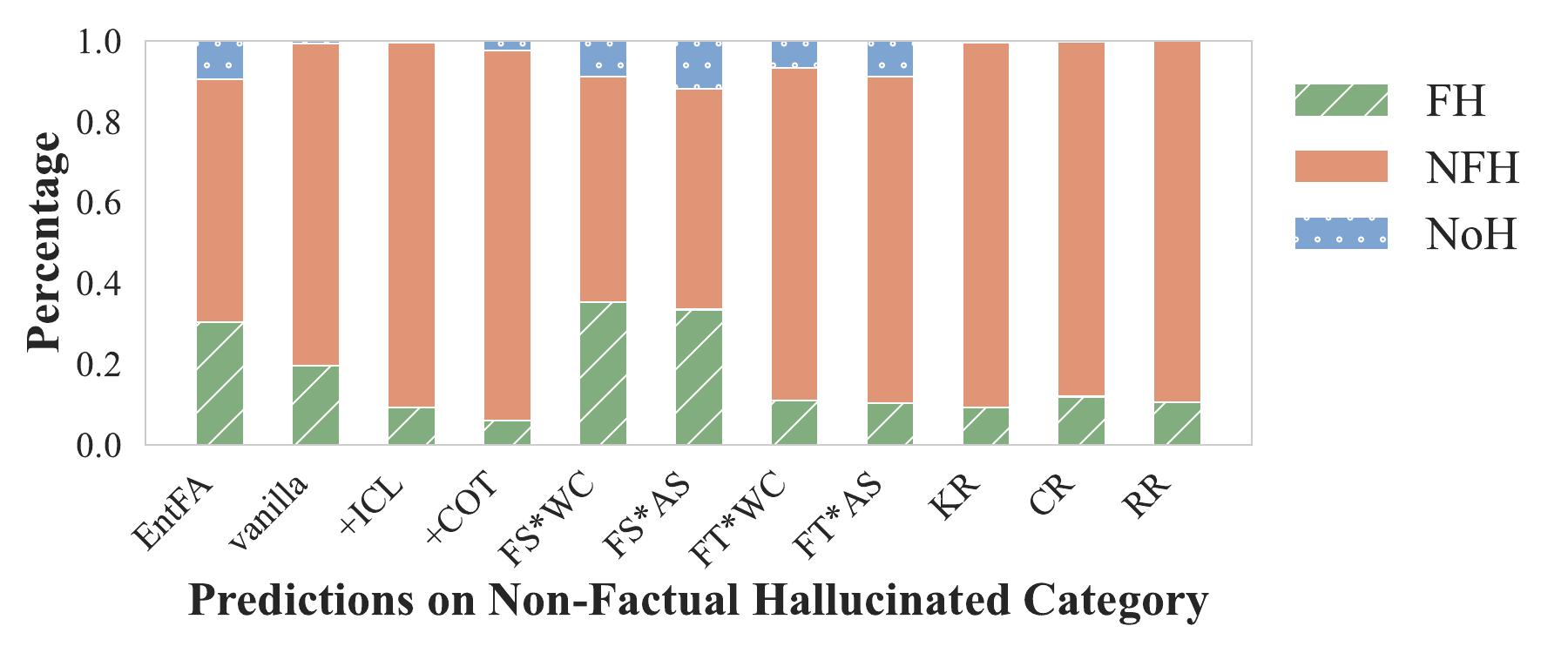} 
        \label{fig:stack_bar_14_2}}
    \subfigure{
        \includegraphics[width=0.47\textwidth]{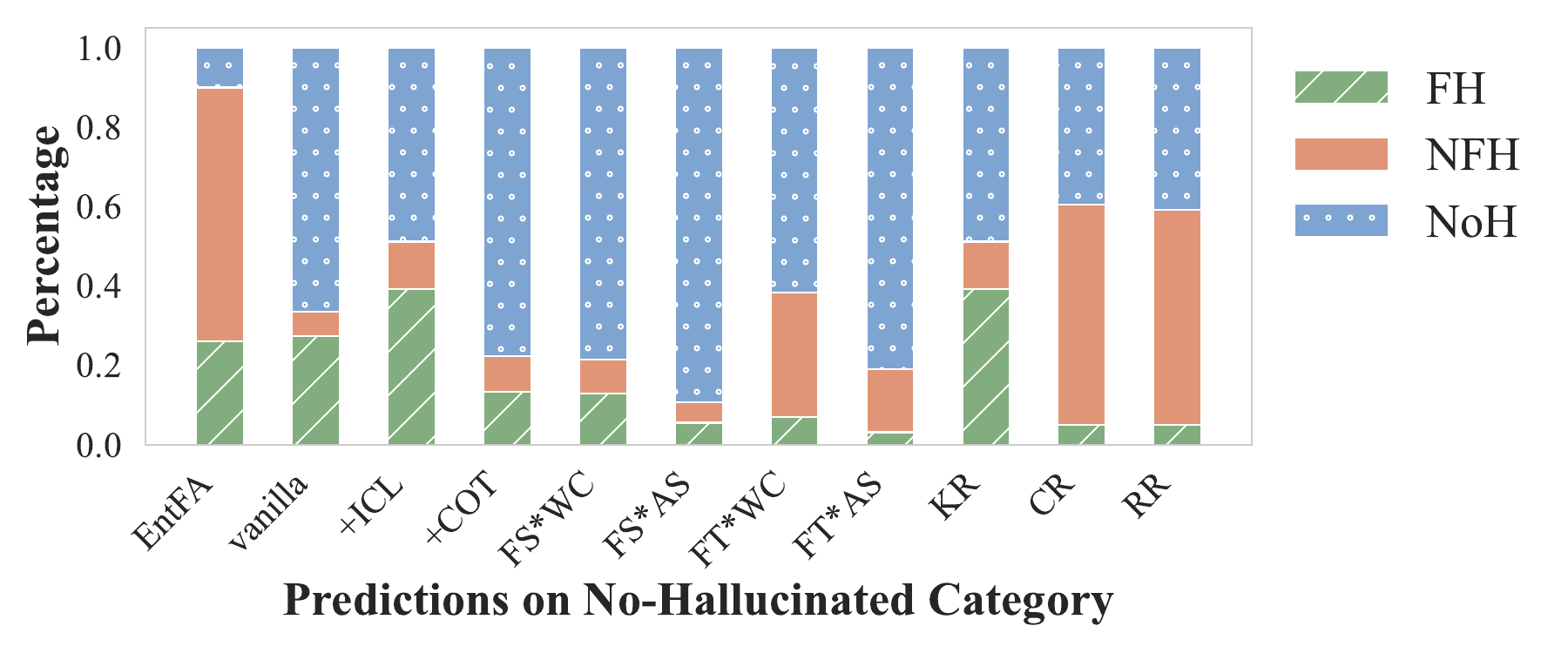}
        \label{fig:stack_bar_14_3}}
    \caption{Predictions of Qwen2.5-14B on categories of FHSumBench.} 
    \label{fig:stack_bar_14}
\end{figure}

\begin{figure}[htbp]
    \centering
    \subfigure{
        \includegraphics[width=0.47\textwidth]{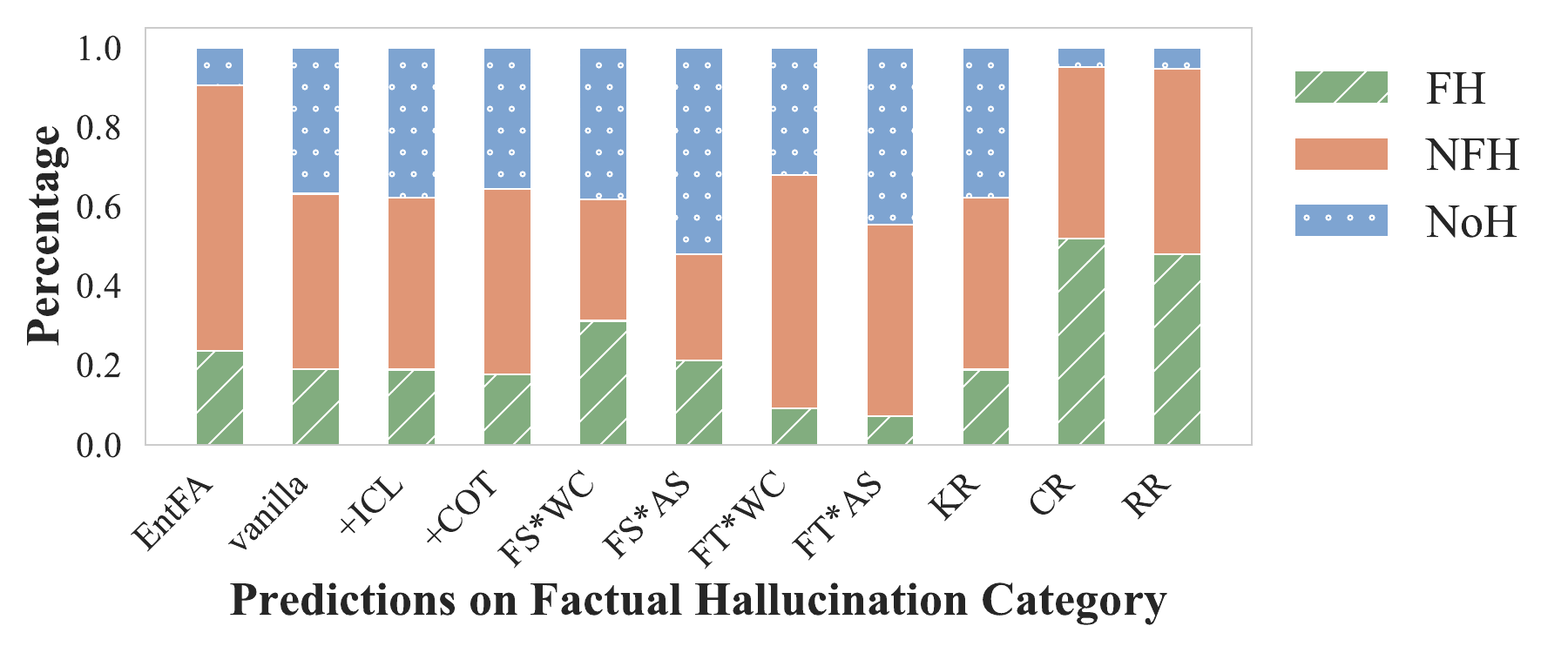} 
        \label{fig:stack_bar_32_1}}
    \subfigure{
        \includegraphics[width=0.47\textwidth]{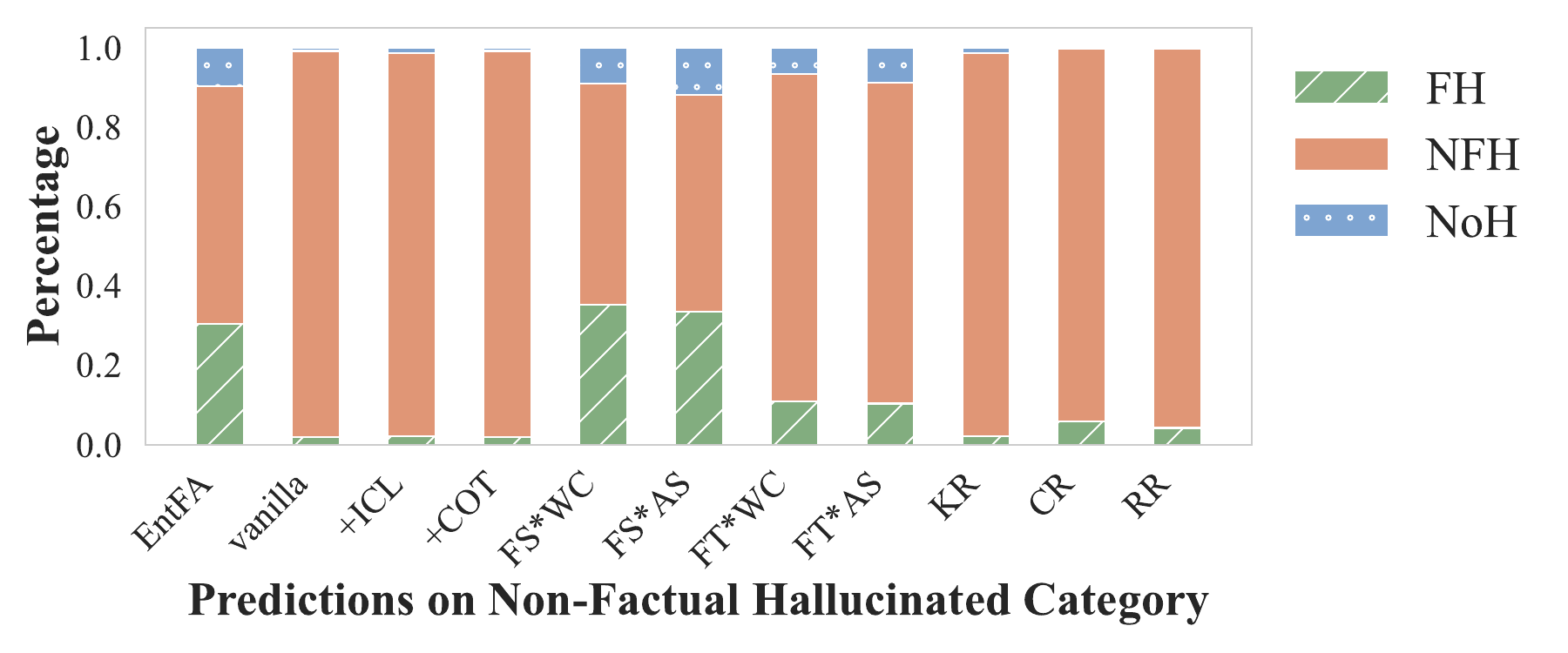} 
        \label{fig:stack_bar_32_2}}
    \subfigure{
        \includegraphics[width=0.47\textwidth]{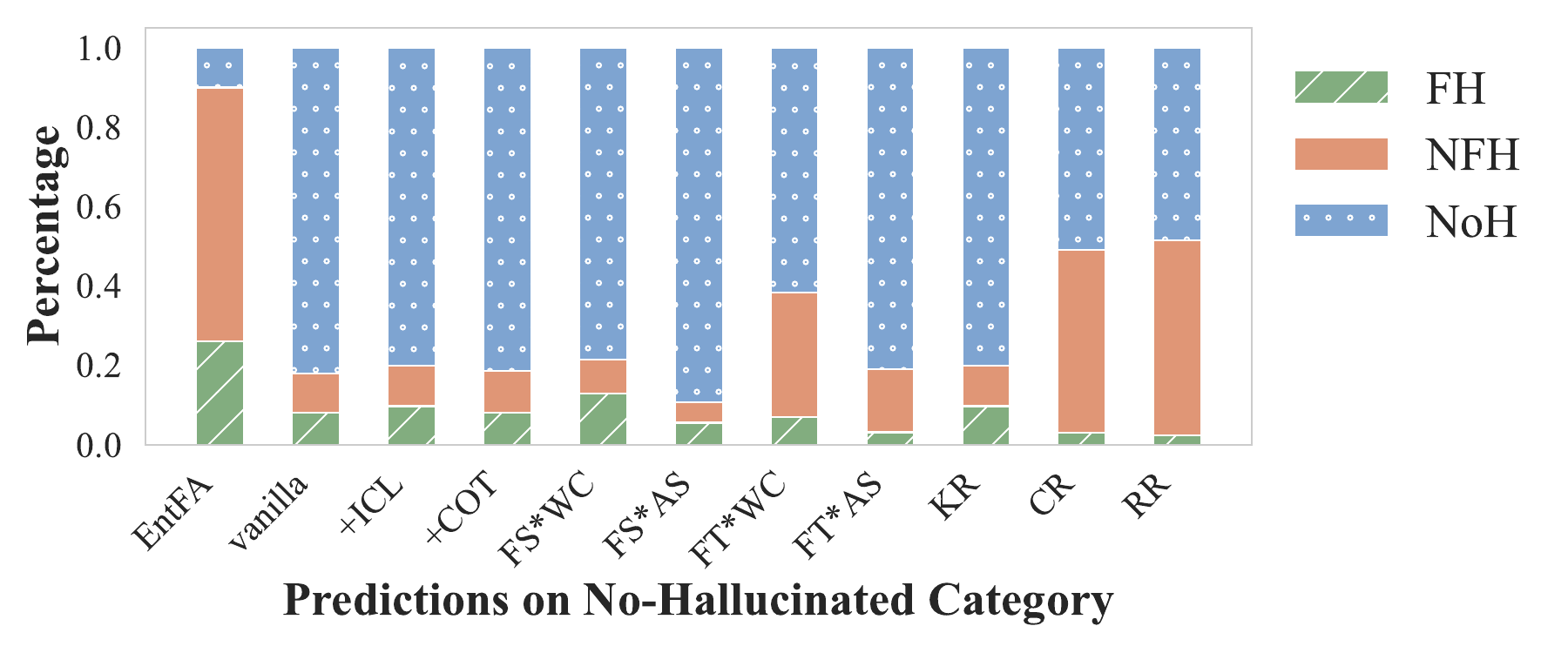}
        \label{fig:stack_bar_32_3}}
    \caption{Predictions of Qwen2.5-32B on categories of FHSumBench.} 
    \label{fig:stack_bar_32}
\end{figure}

Table~\ref{tab:full_categories} shows the full results for categories in FHSumBench and M-XSum, including factual hallucination accuracy (FH-Acc), non-factual hallucination accuracy (NFH-Acc), and no-hallucination accuracy (NoH-Acc). Figure~\ref{fig:stack_bar_8},\ref{fig:stack_bar_14},\ref{fig:stack_bar_32} show the detailed prediction distribution on each category for Llama3-8B, Qwen2.5-14B and Qwen2.5-32B. 
Regarding factual hallucinations, all models demonstrate patterns consistent with GPT-4's results, indicating that CR and RR methods more accurately identify factual hallucinations. For Llama3-8B, all evaluation methods show decreased performance in detecting non-factual hallucinations. Furthermore, larger models like Qwen2.5-32B exhibit more consistent evaluation performance across all methods when assessing non-hallucinated content.
Table~\ref{tab:full_res_scale} presents the full results for LLM-based methods on Qwen2.5 families models, where a darker background color indicates a higher score.

\section{Further Analysis}

\subsection{Case Study Examples}
\label{apx:case_study}
This section provides examples of the case study (\textsection\ref{sec:case}), see Table~\ref{tab:example_case_study}, including the formatted generation of each method.

\subsection{Checking on GPT-4o Generated Knowledge Base}
\label{apx:gpt_knowledge}

In our attempt to integrate entity knowledge from the summaries, we first explored using Wikidata~\cite{10.1145/2629489} as the knowledge base, leveraging entity aliases to expand the entries. However, the coverage proved insufficient, with only 35.9\% of entities in the FHSubBench summaries matching Wikidata entries. This limitation is not unique to Wikidata, when using the Wikipedia database with FactScore, only 55.2\% of entities in FHSumBench and 45\% in M-XSum could be found with entity links in the knowledge graph.
To overcome these coverage limitations, we developed an alternative approach using GPT-4o to generate a comprehensive knowledge base for the summary entities. Specifically, we employ the knowledge evidence generation prompt (detailed in Appendix~\ref{apx:prompts}) to have GPT-4o create Wikipedia-like descriptions for all entities in the text to be evaluated, thereby constructing a knowledge evidence pool.

We have examined the performance of GPT-4o in generating Wikipedia-like knowledge for the entities found in the summaries. After examining 50 samples, we find that 94\% of the generated knowledge is accurate. The remaining 6\% contains errors due to the following reasons:

\begin{itemize}
    \item Contextual ambiguity: The generated entity descriptions sometimes lack contextual information, because entities may have different meanings in different contexts.
    \item Hallucination: For certain entity types such as personal names, organization names, and business names, which may be out of vocabulary, the model occasionally fabricates information.
    \item Timeline issue: For past text content, the generated entity descriptions are based on the current situation. For example, in 2016, Donald Trump had not yet become president, but the generated knowledge states, "Donald Trump is the 45th President of the United States, serving from 2017 to 2021."
\end{itemize}
Despite the above issues, the knowledge generated by LLMs is still more informative than fixed databases like Wikipedia, making it more useful for evaluation.

\clearpage

\begin{table*}[htbp]
    \centering
    \begin{adjustbox}{width=0.7\textwidth}
    \begin{tabular}{cl|cccccc}
    \Xhline{1.5px}
    \multicolumn{2}{c|}{\multirow{3}{*}{Methods}} & \multicolumn{6}{c}{\textbf{FHSumBench}} \\ \hline
    \multicolumn{2}{c|}{} & \multicolumn{1}{c}{\textbf{FH}} & \multicolumn{1}{c}{\textbf{NFH}} & \multicolumn{1}{c}{\textbf{NoH}} & \multicolumn{3}{c}{\textbf{Overall}} \\
    \multicolumn{2}{c|}{} & \multicolumn{1}{c}{\textbf{Acc}} & \multicolumn{1}{c}{\textbf{Acc}} & \multicolumn{1}{c}{\textbf{Acc}} & \multicolumn{1}{c}{\textbf{P}} & \multicolumn{1}{c}{\textbf{R}} & \multicolumn{1}{c}{\textbf{F}} \\ \hline
    \multirow{7}{*}{Vallina} & Qwen2.5-0.5b & \cellcolor{midgreen!14}0.1437 & \cellcolor{midgreen!13}0.1293 & \cellcolor{midgreen!49}0.4941 & \cellcolor{midgreen!26}0.2631 & \cellcolor{midgreen!19}0.1918 & \cellcolor{midgreen!22}0.2219 \\
     & Qwen2.5-1.5b & \cellcolor{midgreen!32}0.3214 & \cellcolor{midgreen!54}0.5415 & \cellcolor{midgreen!54}0.5388 & \cellcolor{midgreen!36}0.3609 & \cellcolor{midgreen!35}0.3504 & \cellcolor{midgreen!36}0.3556 \\
     & Qwen2.5-3b & \cellcolor{midgreen!4}0.0419 & \cellcolor{midgreen!100}1.0000 & \cellcolor{midgreen!1}0.0118 & \cellcolor{midgreen!50}0.5039 & \cellcolor{midgreen!35}0.3512 & \cellcolor{midgreen!41}0.4140 \\
     & Qwen2.5-7b & \cellcolor{midgreen!34}0.3373 & \cellcolor{midgreen!91}0.9122 & \cellcolor{midgreen!63}0.6329 & \cellcolor{midgreen!46}0.4587 & \cellcolor{midgreen!47}0.4706 & \cellcolor{midgreen!46}0.4646 \\
     & Qwen2.5-14b & \cellcolor{midgreen!41}0.4092 & \cellcolor{midgreen!79}0.7902 & \cellcolor{midgreen!66}0.6612 & \cellcolor{midgreen!45}0.4549 & \cellcolor{midgreen!47}0.4652 & \cellcolor{midgreen!46}0.4600 \\
     & Qwen2.5-32b & \cellcolor{midgreen!19}0.1896 & \cellcolor{midgreen!97}0.9707 & \cellcolor{midgreen!82}0.8165 & \cellcolor{midgreen!49}0.4869 & \cellcolor{midgreen!49}0.4942 & \cellcolor{midgreen!49}0.4905 \\
     & Qwen2.5-72b & \cellcolor{midgreen!18}0.1836 & \cellcolor{midgreen!94}0.9415 & \cellcolor{midgreen!76}0.7600 & \cellcolor{midgreen!44}0.4355 & \cellcolor{midgreen!47}0.4713 & \cellcolor{midgreen!45}0.4527 \\ \hline
    \multirow{7}{*}{+ICL} & Qwen2.5-0.5b & \cellcolor{midgreen!15}0.1517 & \cellcolor{midgreen!22}0.2244 & \cellcolor{midgreen!50}0.5035 & \cellcolor{midgreen!27}0.2690 & \cellcolor{midgreen!22}0.2199 & \cellcolor{midgreen!24}0.2420 \\
     & Qwen2.5-1.5b & \cellcolor{midgreen!36}0.3613 & \cellcolor{midgreen!53}0.5293 & \cellcolor{midgreen!41}0.4071 & \cellcolor{midgreen!35}0.3477 & \cellcolor{midgreen!32}0.3244 & \cellcolor{midgreen!34}0.3356 \\
     & Qwen2.5-3b & \cellcolor{midgreen!16}0.1597 & \cellcolor{midgreen!96}0.9561 & \cellcolor{midgreen!9}0.0918 & \cellcolor{midgreen!34}0.3390 & \cellcolor{midgreen!30}0.3019 & \cellcolor{midgreen!32}0.3194 \\
     & Qwen2.5-7b & \cellcolor{midgreen!53}0.5329 & \cellcolor{midgreen!71}0.7146 & \cellcolor{midgreen!49}0.4894 & \cellcolor{midgreen!45}0.4496 & \cellcolor{midgreen!43}0.4342 & \cellcolor{midgreen!44}0.4418 \\
     & Qwen2.5-14b & \cellcolor{midgreen!41}0.4072 & \cellcolor{midgreen!90}0.9000 & \cellcolor{midgreen!49}0.4871 & \cellcolor{midgreen!43}0.4333 & \cellcolor{midgreen!45}0.4486 & \cellcolor{midgreen!44}0.4408 \\
     & Qwen2.5-32b & \cellcolor{midgreen!19}0.1856 & \cellcolor{midgreen!96}0.9610 & \cellcolor{midgreen!80}0.7976 & \cellcolor{midgreen!47}0.4740 & \cellcolor{midgreen!49}0.4861 & \cellcolor{midgreen!48}0.4800 \\
     & Qwen2.5-72b & \cellcolor{midgreen!29}0.2874 & \cellcolor{midgreen!85}0.8512 & \cellcolor{midgreen!69}0.6871 & \cellcolor{midgreen!43}0.4311 & \cellcolor{midgreen!46}0.4564 & \cellcolor{midgreen!44}0.4434 \\ \hline
    \multirow{7}{*}{+COT} & Qwen2.5-0.5b & \cellcolor{midgreen!22}0.2196 & \cellcolor{midgreen!21}0.2098 & \cellcolor{midgreen!42}0.4188 & \cellcolor{midgreen!26}0.2643 & \cellcolor{midgreen!21}0.2120 & \cellcolor{midgreen!24}0.2353 \\
     & Qwen2.5-1.5b & \cellcolor{midgreen!11}0.1098 & \cellcolor{midgreen!90}0.9049 & \cellcolor{midgreen!5}0.0471 & \cellcolor{midgreen!37}0.3704 & \cellcolor{midgreen!27}0.2654 & \cellcolor{midgreen!31}0.3092 \\
     & Qwen2.5-3b & \cellcolor{midgreen!11}0.1098 & \cellcolor{midgreen!90}0.9049 & \cellcolor{midgreen!5}0.0471 & \cellcolor{midgreen!37}0.3704 & \cellcolor{midgreen!27}0.2654 & \cellcolor{midgreen!31}0.3092 \\
     & Qwen2.5-7b & \cellcolor{midgreen!31}0.3054 & \cellcolor{midgreen!91}0.9073 & \cellcolor{midgreen!46}0.4612 & \cellcolor{midgreen!43}0.4345 & \cellcolor{midgreen!42}0.4185 & \cellcolor{midgreen!43}0.4263 \\
     & Qwen2.5-14b & \cellcolor{midgreen!26}0.2595 & \cellcolor{midgreen!88}0.8756 & \cellcolor{midgreen!76}0.7647 & \cellcolor{midgreen!47}0.4717 & \cellcolor{midgreen!47}0.4749 & \cellcolor{midgreen!47}0.4733 \\
     & Qwen2.5-32b & \cellcolor{midgreen!18}0.1756 & \cellcolor{midgreen!96}0.9634 & \cellcolor{midgreen!81}0.8094 & \cellcolor{midgreen!48}0.4808 & \cellcolor{midgreen!49}0.4871 & \cellcolor{midgreen!48}0.4840 \\
     & Qwen2.5-72b & \cellcolor{midgreen!13}0.1257 & \cellcolor{midgreen!91}0.9098 & \cellcolor{midgreen!84}0.8400 & \cellcolor{midgreen!48}0.4801 & \cellcolor{midgreen!47}0.4689 & \cellcolor{midgreen!47}0.4744 \\ \hline
    \multirow{7}{*}{KR} & Qwen2.5-0.5b & \cellcolor{midgreen!12}0.1178 & \cellcolor{midgreen!9}0.0902 & \cellcolor{midgreen!49}0.4871 & \cellcolor{midgreen!24}0.2427 & \cellcolor{midgreen!17}0.1738 & \cellcolor{midgreen!20}0.2025 \\
     & Qwen2.5-1.5b & \cellcolor{midgreen!27}0.2655 & \cellcolor{midgreen!51}0.5098 & \cellcolor{midgreen!57}0.5741 & \cellcolor{midgreen!34}0.3452 & \cellcolor{midgreen!34}0.3373 & \cellcolor{midgreen!34}0.3412 \\
     & Qwen2.5-3b & \cellcolor{midgreen!1}0.0120 & \cellcolor{midgreen!100}1.0000 & \cellcolor{midgreen!1}0.0094 & \cellcolor{midgreen!56}0.5622 & \cellcolor{midgreen!34}0.3405 & \cellcolor{midgreen!42}0.4241 \\
     & Qwen2.5-7b & \cellcolor{midgreen!21}0.2056 & \cellcolor{midgreen!92}0.9244 & \cellcolor{midgreen!61}0.6094 & \cellcolor{midgreen!42}0.4182 & \cellcolor{midgreen!43}0.4348 & \cellcolor{midgreen!43}0.4263 \\
     & Qwen2.5-14b & \cellcolor{midgreen!32}0.3174 & \cellcolor{midgreen!92}0.9244 & \cellcolor{midgreen!65}0.6541 & \cellcolor{midgreen!46}0.4628 & \cellcolor{midgreen!47}0.4740 & \cellcolor{midgreen!47}0.4683 \\
     & Qwen2.5-32b & \cellcolor{midgreen!13}0.1297 & \cellcolor{midgreen!99}0.9878 & \cellcolor{midgreen!80}0.8047 & \cellcolor{midgreen!48}0.4750 & \cellcolor{midgreen!48}0.4806 & \cellcolor{midgreen!48}0.4777 \\
     & Qwen2.5-72b & \cellcolor{midgreen!13}0.1297 & \cellcolor{midgreen!96}0.9659 & \cellcolor{midgreen!81}0.8071 & \cellcolor{midgreen!45}0.4549 & \cellcolor{midgreen!48}0.4757 & \cellcolor{midgreen!47}0.4650 \\ \hline
    \multirow{7}{*}{CR} & Qwen2.5-0.5b & \cellcolor{midgreen!34}0.3433 & \cellcolor{midgreen!69}0.6927 & \cellcolor{midgreen!1}0.0047 & \cellcolor{midgreen!44}0.4425 & \cellcolor{midgreen!26}0.2602 & \cellcolor{midgreen!33}0.3277 \\
     & Qwen2.5-1.5b & \cellcolor{midgreen!6}0.0579 & \cellcolor{midgreen!98}0.9854 & \cellcolor{midgreen!1}0.0118 & \cellcolor{midgreen!71}0.7134 & \cellcolor{midgreen!35}0.3517 & \cellcolor{midgreen!47}0.4711 \\
     & Qwen2.5-3b & \cellcolor{midgreen!49}0.4870 & \cellcolor{midgreen!74}0.7439 & \cellcolor{midgreen!18}0.1765 & \cellcolor{midgreen!51}0.5085 & \cellcolor{midgreen!35}0.3518 & \cellcolor{midgreen!42}0.4159 \\
     & Qwen2.5-7b & \cellcolor{midgreen!59}0.5948 & \cellcolor{midgreen!71}0.7122 & \cellcolor{midgreen!44}0.4353 & \cellcolor{midgreen!54}0.5362 & \cellcolor{midgreen!44}0.4356 & \cellcolor{midgreen!48}0.4807 \\
     & Qwen2.5-14b & \cellcolor{midgreen!50}0.4970 & \cellcolor{midgreen!79}0.7927 & \cellcolor{midgreen!37}0.3694 & \cellcolor{midgreen!53}0.5325 & \cellcolor{midgreen!41}0.4148 & \cellcolor{midgreen!47}0.4663 \\
     & Qwen2.5-32b & \cellcolor{midgreen!46}0.4551 & \cellcolor{midgreen!87}0.8683 & \cellcolor{midgreen!48}0.4776 & \cellcolor{midgreen!57}0.5650 & \cellcolor{midgreen!45}0.4503 & \cellcolor{midgreen!50}0.5012 \\
     & Qwen2.5-72b & \cellcolor{midgreen!51}0.5110 & \cellcolor{midgreen!80}0.7976 & \cellcolor{midgreen!50}0.4988 & \cellcolor{midgreen!56}0.5596 & \cellcolor{midgreen!45}0.4518 & \cellcolor{midgreen!50}0.5000 \\ \hline
    \multirow{7}{*}{RR} & Qwen2.5-0.5b & \cellcolor{midgreen!49}0.4870 & \cellcolor{midgreen!64}0.6415 & \cellcolor{midgreen!1}0.0071 & \cellcolor{midgreen!46}0.4598 & \cellcolor{midgreen!28}0.2839 & \cellcolor{midgreen!35}0.3510 \\
     & Qwen2.5-1.5b & \cellcolor{midgreen!5}0.0479 & \cellcolor{midgreen!97}0.9780 & \cellcolor{midgreen!1}0.0047 & \cellcolor{midgreen!49}0.4895 & \cellcolor{midgreen!26}0.2577 & \cellcolor{midgreen!34}0.3376 \\
     & Qwen2.5-3b & \cellcolor{midgreen!50}0.4950 & \cellcolor{midgreen!73}0.7366 & \cellcolor{midgreen!20}0.2000 & \cellcolor{midgreen!50}0.4970 & \cellcolor{midgreen!36}0.3579 & \cellcolor{midgreen!42}0.4161 \\
     & Qwen2.5-7b & \cellcolor{midgreen!57}0.5709 & \cellcolor{midgreen!73}0.7341 & \cellcolor{midgreen!44}0.4424 & \cellcolor{midgreen!55}0.5470 & \cellcolor{midgreen!44}0.4368 & \cellcolor{midgreen!49}0.4858 \\
     & Qwen2.5-14b & \cellcolor{midgreen!47}0.4711 & \cellcolor{midgreen!81}0.8146 & \cellcolor{midgreen!38}0.3812 & \cellcolor{midgreen!54}0.5395 & \cellcolor{midgreen!42}0.4167 & \cellcolor{midgreen!47}0.4702 \\
     & Qwen2.5-32b & \cellcolor{midgreen!46}0.4251 & \cellcolor{midgreen!87}0.8902 & \cellcolor{midgreen!48}0.4541 & \cellcolor{midgreen!57}0.5648 & \cellcolor{midgreen!45}0.4424 & \cellcolor{midgreen!50}0.4961 \\
     & Qwen2.5-72b & \cellcolor{midgreen!51}0.4810 & \cellcolor{midgreen!80}0.8073 & \cellcolor{midgreen!50}0.5153 & \cellcolor{midgreen!56}0.5666 & \cellcolor{midgreen!45}0.4509 & \cellcolor{midgreen!50}0.5022 \\  \Xhline{1.5px}
    \end{tabular}
    \end{adjustbox}
    \caption{Full results for LLM-based methods on Qwen2.5 families models.}
    \label{tab:full_res_scale}
    \end{table*}

\newcommand{\jsonindent}{\hspace{1em}}
\onecolumn
\begin{longtable}{p{2cm}p{\dimexpr\linewidth-3cm-4\tabcolsep}}
\hline \hline
\textbf{Text} & \textbf{Content} \\
\midrule
\endfirsthead

\multicolumn{2}{l}{Continued from previous page...}\\
\toprule
\textbf{Method} & \textbf{Response} \\
\midrule
\endhead

\midrule
\multicolumn{2}{r}{Continued on next page...}\\
\endfoot

\bottomrule
\caption{Case Study.}
\label{tab:example_case_study}
\endlastfoot

Document & Those who participated in the Aberdeen Children of the 1950s project, which saw all primary pupils aged seven to 12 surveyed by the Medical Research Council in 1962, have been contacted. They have been asked to take part in the Scottish Family Health Study, which is recruiting 50,000 people. It aims to investigate why diseases such as cancer can run in families. Those recruited will have their health tracked, with the intention of creating a Scottish \"bio-bank\" containing genetic, medical and family history and lifestyle information. This will allow researchers to investigate the contribution of genetic and environmental factors to common conditions. Blair Smith, a GP and professor of primary care medicine at Aberdeen University, will run the project. He said: \"We know that many diseases such as cancer, heart disease, mental illness and diabetes run in families and that tells us that genes are important. \"But we don't know yet which genes are important or how they relate to other things such as lifestyle. \"We can get more information about genes by studying DNA which carries the genetic blueprint handed down through the generations. \"That is why we are collecting and analysing blood samples and health information from thousands of families across Scotland.\" He said the data gathered would help future research into the prevention, treatment and diagnosis of illnesses. \\ \hline
Summary & Aberdeen children from the 1950s have been invited to take part in a major study of family health in Scotland, home nation of the United Kingdom, as part of a new project. \\ \hline
Label & Factual hallucination ("home nation of the United Kingdom" is a correct description of "Scotland"). \\ \hline \hline
\rule{0pt}{10pt} \textbf{Method} & \textbf{Response} \\ \hline

Vanilla Judge & \{\par\jsonindent"Faithfulness": false, \par \jsonindent"Factuality": false\par\} \\ \hline
+ICL &\{\par\jsonindent"Faithfulness": true, \par \jsonindent"Factuality": true\par\} \\ 
+CoT & To evaluate the faithfulness and factuality of the summary in relation to the document, let's first examine the key points from each:\par 1. **Document Overview**: \par - Participants from the Aberdeen Children of the 1950s project are invited to the Scottish Family Health Study.  - The study aims to recruit 50,000 people to investigate hereditary diseases.\par - The study will create a  "bio-bank "including genetic, medical, history, and lifestyle information.  - Led by GP and professor Blair Smith, the project will analyze DNA to understand genetic factors in diseases.\par - The broader aim is to aid future research in disease prevention, treatment, and diagnosis.\par     \par 2. **Summary Overview**:\par  - States that Aberdeen children from the 1950s are invited to join a family health study in Scotland.\par \par 3. **Analysis**:\par - **Faithfulness**:\par  - The summary accurately conveys that individuals from the Aberdeen Children of the 1950s project are being contacted for a new study.\par  - It specifically highlights the location (Scotland) and timing (from the 1950s project).\par  - It does not mention the scope, goals, leader, or the extensive nature of the Scottish Family Health Study, which indicates missing information but not necessarily deviation or misrepresentation from what's present.\par \par - **Factuality**:\par  - The statement regarding Aberdeen children from the 1950s and their involvement in a new project aligns with the document.\par  - The phrase  "home nation of the United Kingdom " is factual but irrelevant to the core details of the study, as the invitation pertains to a Scottish study regardless of broader geopolitical context.\par \par **Conclusion**:  The summary does not contain any statements that deviate from or misrepresent the document (it is faithful to what it covers, but lacks completeness). It also does not present any factual inaccuracies; it simply omits details present in the document. Therefore, based on what's provided, the summary is both faithful and factual.\par  \{\par\jsonindent"Faithfulness": true, \par \jsonindent"Factuality": true\par\} " \\ \hline
Knowledge Retrieve & \{\par\jsonindent"Faithfulness": true, \par \jsonindent"Factuality": true\par\} \\ 
Concurrent Retrieve & queries: \par \jsonindent1. "1950s Aberdeen children invited major study 2023",\par \jsonindent2. "family health study Scotland site:.gov OR site:.ac.uk",\par \jsonindent 3. "Scotland home nation United Kingdom official status",\par \jsonindent 4."new project study evidence 2023 site:.edu”\par \par judgement:\par \jsonindent 1. claim: Aberdeen children from the 1950s have been invited to a major study., judge: faithful\par \jsonindent 2. claim: The study is about family health in Scotland., judge: faituful\par \jsonindent 3. claim: Scotland is a home nation of the United Kingdom., judge: factual hallucination\par \jsonindent 4. claim: The study is part of a new project., judge: unknown \\ \hline
Reflection Retrieve & queries:\par \jsonindent 1. "1950s Aberdeen children invited major study 2023",\par \jsonindent 2. "study on family health outcomes in Scotland 2023",\par \jsonindent 3. "Scotland status as home nation of United Kingdom evidence", (reflective query: "Scotlands political status within the United Kingdom")\par \jsonindent 4. "study is part of a new project " evidence 2023 academic research publication"\par \par judgement:\par \jsonindent 1. claim: Aberdeen children from the 1950s have been invited to a major study.,  judge: faithful\par \jsonindent 2. claim: The study is about family health in Scotland., judge: faithful\par \jsonindent 3. claim: Scotland is a home nation of the United Kingdom., judge: factual hallucination\par \jsonindent 4. claim: The study is part of a new project.,  judge: faithful \\ 
\end{longtable}

\clearpage

\section{Prompts}
\label{apx:prompts}
The following are the prompts used in the experiments.

\begin{tcolorbox}[
    enhanced,
    attach boxed title to top left={
        yshift=-3mm,yshifttext=-1mm
    },
    breakable,
    colframe=gray!75!black,
    colbacktitle=gray,
    title=Vanilla Judge Prompt,
    coltitle=black,
    collower=purple, 
    label=vanilla,
    fonttitle=\bfseries,
    boxed title style={size=small,colframe=gray!50!black, colback=light_grey}
    ]
    Task Description:
    You are tasked with assessing the hallucination of a summary for a given document. Your role is to evaluate whether the summary: \\

    1. Contains any information that deviates from or misrepresents the content of the document (Faithfulness). \\
    2. Includes factual inaccuracies based on common knowledge or common sense (Factuality). \\

    You will have access to the document to assist in your judgment. Provide your judgment in the following JSON format: \\
    \{ \\
    "Faithfulness": true/false, \\
    "Factuality": true/false \\
    \} \\

    Definitions: \\
    1. Faithfulness: Determines if the information in the summary is directly inferred or logically entailed from the content of the document. \\
    2. Factuality: Assesses whether the information in the summary aligns with commonly accepted facts, general knowledge, or common sense. \\

    You should try your best to determine if the summary contains non-factual or hallucinated information according to the above hallucination types. Always provide answers in JSON format, start with "\{", adhering strictly to the schema provided above. Do not include any explanations or extra content. \\

    Document:\{doc\} \\
    Summary:\{summary\} \\
    Your Judgement:

\end{tcolorbox}

\clearpage
\begin{tcolorbox}[
    enhanced,
    attach boxed title to top left={
        yshift=-3mm,yshifttext=-1mm
    },
    breakable,
    colframe=gray!75!black,
    colbacktitle=gray,
    title=ICL Judge Prompt (3-shot),
    coltitle=black,
    collower=purple, 
    label=icl,
    fonttitle=\bfseries,
    boxed title style={size=small,colframe=gray!50!black, colback=light_grey}
    ]
    (... previous task description omitted ...) \\

    Example 1:

    Document: Two leading groups, Jaysh al-Islam and Ahrar al-Sham, which formed a pact last year, both say the plane was shot down. Syrian state media said the crash was caused by a technical fault...... 

    Your Judgement: \\
    \{ \\
    "Faithfulness": false, \\
    "Factuality": true \\
    \} \\

    Example 2:
        
    Document: These are external links and will open in a new window. The units in North Tyneside and Northumberland have been shut between midnight and 08:00 since December. Overnight emergencies have been diverted to the recently-opened Northumbria Hospital in Cramlington......

    Your Judgement: \\
    \{ \\
    "Faithfulness": true, \\
    "Factuality": true \\
    \} \\

    Example 3:
        
    Document: In a damning new report, the group also called for an \"independent and impartial\" inquiry into cases of abuse. The law, AFSPA, was introduced in the region in 1990 as a response to violence by insurgent groups......

    Your Judgement: \\
    \{ \\
    "Faithfulness": false, \\
    "Factuality": false \\
    \} \\

    (... following task description omitted ...) \\

\end{tcolorbox}

\begin{tcolorbox}[
    enhanced,
    attach boxed title to top left={
        yshift=-3mm,yshifttext=-1mm
    },
    breakable,
    colframe=gray!75!black,
    colbacktitle=gray,
    title=CoT Judge Prompt,
    coltitle=black,
    collower=purple, 
    label=cot,
    fonttitle=\bfseries,
    boxed title style={size=small,colframe=gray!50!black, colback=light_grey}
    ]
    (... previous task description omitted ...) \\

    You should try your best to determine if the summary contains non-factual or hallucinated information according to the above hallucination types. Think step by step, and provide the trajectory before your judgement within 200 tokens. Always provide your final judgement in JSON format, start with "\{", adhering strictly to the schema provided above.

\end{tcolorbox}

Below are the prompts used in retrieval-based methods.
\clearpage
\begin{tcolorbox}[
    enhanced,
    attach boxed title to top left={
        yshift=-3mm,yshifttext=-1mm
    },
    breakable,
    colframe=yellow!75!black,
    colbacktitle=yellow,
    title=Claim Extraction Prompt,
    coltitle=black,
    collower=purple, 
    label=retrieve_claim,
    fonttitle=\bfseries,
    boxed title style={size=small,colframe=yellow!50!black, colback=yellow}
    ]
    Task description:
    Break the given text into several independent claims, resolve the coreference. A claim is a statement that represents a clear and self-contained fact, opinion, or assertion, which can be verified by humans. Your task is to accurately identify and extract every claim stated in the provided text. Then, resolve any coreference (pronouns or other referring expressions) in the claim for clarity. Each claim should be concise (less than 15 words)Each claim should represent a clear and self-contained fact, opinion, or assertion. 
    Split the claims with "\textbackslash n". Start with the claims. Do not omit any information in the sentence. DO NOT RESPOND WITH ANYTHING ELSE. \\

    Example 1:  \\
        Sentence: "This company offers high-quality products, and its customer service is highly regarded in the industry." \\
        Claims:  \\
            This company offers high-quality products. \\
            This company has customer service. \\
            The customer service is highly regarded in the industry. \\
    
    Example 2: \\
        Sentence: "Former Wales captain Martyn Williams says Dan Biggar's decision to sign a new contract with Ospreys will benefit the region." \\
        Claims: \\
            Martyn Williams is a former Wales captain. \\
            Martyn Williams says about Dan Biggar. \\
            Dan Biggar has decided to sign a new contract with Ospreys \\
            Martyn Williams believes this decision will benefit the Ospreys region. \\

    Example 3: \\
        Sentence: "This city not only has a rich cultural heritage but is also an important economic center that attracts a lot of investment." \\
        Claims: \\
            This city has a rich cultural heritage. \\
            This city is an important economic center. \\
            This important economic center attracts a lot of investment. \\
    
    Now, please break down the following sentence: \\
        Sentence: \{summary\} \\
        Claims:

\end{tcolorbox}

\clearpage

\begin{tcolorbox}[
    enhanced,
    attach boxed title to top left={
        yshift=-3mm,yshifttext=-1mm
    },
    breakable,
    colframe=yellow!75!black,
    colbacktitle=yellow,
    title=Knowledge Evidence Generation Prompt,
    coltitle=black,
    collower=purple, 
    label=knowledge_generation,
    fonttitle=\bfseries,
    boxed title style={size=small,colframe=yellow!50!black, colback=yellow}
    ]
    Describe the following entity in a concise and informative manner. Include key characteristics, functions, and any relevant context that helps explain its significance or role. Aim for a clear and engaging description within a few sentences. Please only reply with the description, DO NOT include any extra content. \\

    Entity: \{entity\} \\
    Description:

\end{tcolorbox}

\begin{tcolorbox}[
    enhanced,
    attach boxed title to top left={
        yshift=-3mm,yshifttext=-1mm
    },
    breakable,
    colframe=yellow!75!black,
    colbacktitle=yellow,
    title=Reflection on Evidence Prompt,
    coltitle=black,
    collower=purple, 
    label=retrieve_reflect,
    fonttitle=\bfseries,
    boxed title style={size=small,colframe=yellow!50!black, colback=yellow}
    ]
    What key information is missing from our current evidence to make a judgement about this claim? Please reply the missing information in the format of a sentence within 20 tokens. If there are multiple missing information, reply the most important one. DO NOT reply any other information. \\

    Claim: \{claim\} \\
    Current Evidence: \{retrieved evidence list\} \\
    Missing Information:

\end{tcolorbox}

\clearpage

\begin{tcolorbox}[
    enhanced,
    attach boxed title to top left={
        yshift=-3mm,yshifttext=-1mm
    },
    breakable,
    colframe=yellow!75!black,
    colbacktitle=yellow,
    title=Judge on Evidence Prompt,
    coltitle=black,
    collower=purple, 
    label=retrieve_judge,
    fonttitle=\bfseries,
    boxed title style={size=small,colframe=yellow!50!black, colback=yellow}
    ]
    Analyze how the evidence relates to the claim. DO NOT make any assumption, reasoning or use previously owned knowledge EXCEPT the evidence. Start with the answer, DO NOT reply any other information.
    Answer in the following format, split with "\textbackslash n": \\
        Rationale: Give the rationale for the answer, within 30 tokens. Relevant means whether the evidence provides connection to the claim or contains the information of the claim. If Relevant is Yes, then make a judgement on relation. \\
        Relevant: Yes / No \\
        Relation: Support / Contradict \\

     Example1: \\
        Claim: New York is an eastern state in the United States of America. \\
        Evidence: New York is known for its iconic landmarks like the Statue of Liberty and bustling New York City. \\
        Answer: \\
            Rationale: Even though the evidence is about New York, it does not directly support the claim that "New York is an eastern state in the United States of America." Instead, it mentions iconic landmarks and features of New York, such as the Statue of Liberty and New York City, without addressing its geographical location or status as an eastern state. Therefore, the evidence is not relevant to the claim. \\
            Relevant: No \\
            Relation: None \\

    Example2: \\
        Claim: Regular exercise improves mental health. \\
        Evidence: Studies show that individuals who engage in physical activity report lower levels of stress and anxiety. \\
        Answer: \\
            Rationale: The evidence indicates a positive relationship between physical activity and reduced mental health issues, aligning with the claim. \\
            Relevant: Yes \\
            Relation: Support \\

    Example3: \\
        Claim: the pilot of the crashed jet killed himself. \\
        Evidence: Jaysh al-Islam, the larger of the two groups, posted footage online which it claimed showed the pilot being held after ejecting from the jet. The video, bearing Jaysh al-Islam's logo, showed an object engulfed in flames followed by an interview with the supposed pilot. \\
        Answer: \\
            Rationale: The evidence addresses the topic of the pilot living status. The evidence suggests the pilot survived and was captured, which is inconsistent with the claim of suicide. \\
            Relevant: Yes \\
            Relation: Contradict \\

    Now, please answer based on the claim and evidence: \\
        Claim: \{claim\} \\
        Evidence: \{candidate\} \\
        Answer: 

\end{tcolorbox}

\end{document}